%% file: Mono3DPose.tex
\documentclass[10pt,twocolumn,letterpaper]{article}

\usepackage{iccv}
\usepackage{times}
\usepackage{epsfig}
\usepackage{graphicx}
\usepackage{amsmath}
\usepackage{amssymb}
\input{macro}

\usepackage[pagebackref=true,breaklinks=true,letterpaper=true,colorlinks,bookmarks=false]{hyperref}

\iccvfinalcopy 

\def\iccvPaperID{423} 

\ificcvfinal\fi
\begin{document}

\title{Generalizing Monocular 3D Human Pose Estimation in the Wild}

\author{Luyang Wang$^{1,2}$\thanks{Indicates equal contribution.} \quad Yan Chen$^{1,2*}$ \quad Zhenhua Guo$^{2}$ \quad Keyuan Qian$^{2}$ \\ Mude Lin$^{1}$ \quad Hongsheng Li$^{3}$\quad Jimmy S. Ren$^{1}$
\\ $^1$SenseTime Research \quad $^2$Tsinghua University \quad $^3$CUHK-SenseTime Joint Lab\\
\{wang-ly16, yan-chen16\}@mails.tsinghua.edu.cn}

\maketitle

\begin{abstract}
The availability of the large-scale labeled 3D poses in the Human3.6M dataset plays an important role in advancing the algorithms for 3D human pose estimation from a still image. 
We observe that recent innovation in this area mainly focuses on new techniques that explicitly address the generalization issue when using this dataset, because this database is constructed in a highly controlled environment with limited human subjects and background variations.
Despite such efforts, we can show that the results of the current methods are still error-prone especially when tested against the images taken in-the-wild. 
In this paper, we aim to tackle this problem from a different perspective. 
We propose a principled approach to generate high quality 3D pose ground truth given any in-the-wild image with a person inside. We achieve this by first devising a novel stereo inspired neural network to directly map any 2D pose to high quality 3D counterpart. 
We then perform a carefully designed geometric searching scheme to further refine the joints. 
Based on this scheme, we build a large-scale dataset with 400,000 in-the-wild images and their corresponding 3D pose ground truth. 
This enables the training of a high quality neural network model, without specialized training scheme and auxiliary loss function, which performs favorably against the state-of-the-art 3D pose estimation methods. 
We also evaluate the generalization ability of our model both quantitatively and qualitatively. 
Results show that our approach convincingly outperforms the previous methods. 
We make our dataset and code publicly available.\footnote{\url{https://github.com/llcshappy/Monocular-3D-Human-Pose}}
\end{abstract}

\section{Introduction}
\begin{figure}[!t]
\vspace{2mm}
\footnotesize
\centering
\renewcommand{\tabcolsep}{1pt} 
\renewcommand{\arraystretch}{1} 
\begin{center}
\resizebox{\linewidth}{!}{%
\begin{tabular}{cccc}
  \includegraphics[width=0.24\linewidth]{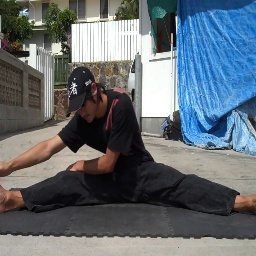} &
  \includegraphics[width=0.24\linewidth]{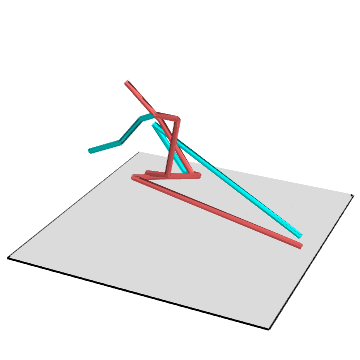} &
  \includegraphics[width=0.24\linewidth]{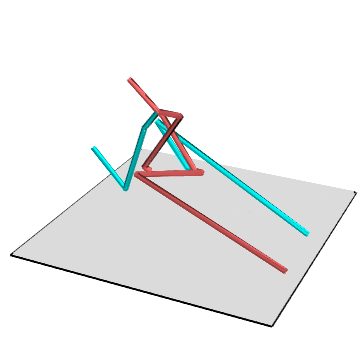} &
  \includegraphics[width=0.24\linewidth]{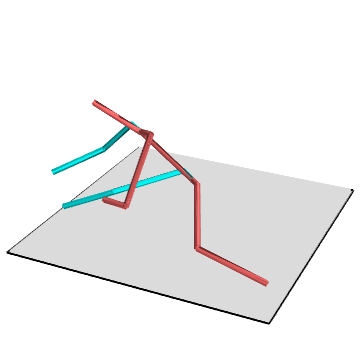} \\
  \includegraphics[width=0.24\linewidth]{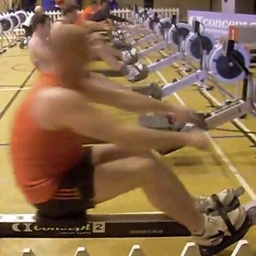} &
  \includegraphics[width=0.24\linewidth]{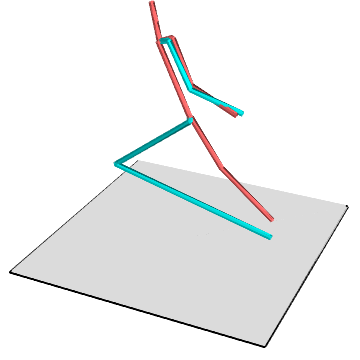} &
  \includegraphics[width=0.24\linewidth]{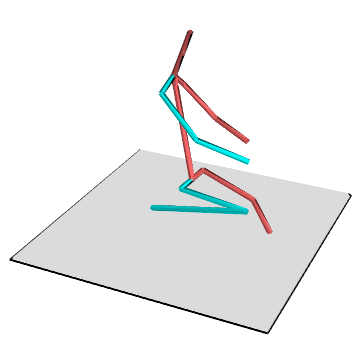} &
  \includegraphics[width=0.24\linewidth]{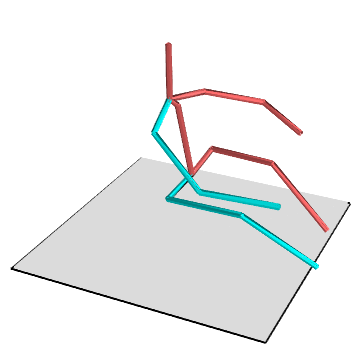} \\
  \includegraphics[width=0.24\linewidth]{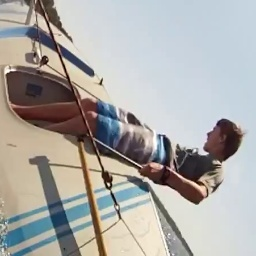} &
  \includegraphics[width=0.24\linewidth]{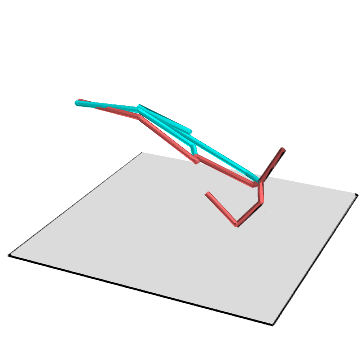} &
  \includegraphics[width=0.24\linewidth]{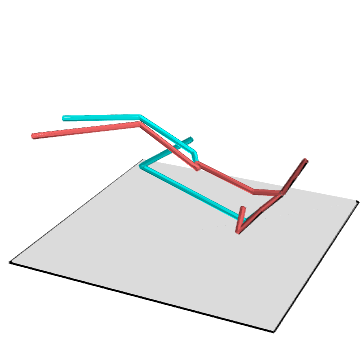} &
  \includegraphics[width=0.24\linewidth]{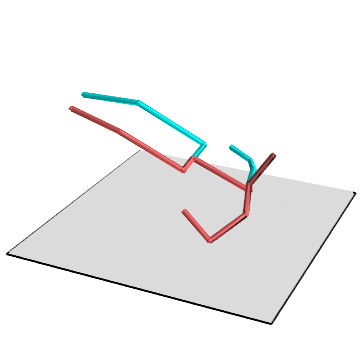} \\
  Original&
  Zhou~\etal~\cite{zhou2017towards}&
  Yang~\etal~\cite{yang20183d}& 
  Ours\\
  \end{tabular}}
\end{center}
\vspace{-2mm}
\caption{3D pose estimation on challenging images.
The proposed method performs favorably against the state-of-the-art 3D pose estimation algorithms due to our generated in-the-wild 3D pose dataset.
The red lines denote the skeletons of the left and torso part of the human body while blue lines represent the right part.
All the predicted poses are demonstrated at the same viewpoint.}
\vspace{-6mm}
\label{intro:3D skeletion}
\end{figure}
3D human pose estimation is one of the fundamental problems in computer vision.
It is widely used in a large number of areas such as action recognition, virtual reality, human-computer interaction, and video surveillance.
Recently, significant advances have been achieved in 2D human pose estimation due to the powerful deep Convolutional Neural Networks (CNNs) and the availability of large-scale in-the-wild 2D human pose datasets with manual annotations.
However, advances in 3D human pose estimation remain limited.
%

This problem is widely studied in the literature and is mainly tackled with the following types of technical methodologies namely 2D-to-3D pose estimation~\cite{fang2017learning, martinez2017simple}, monocular image based 3D pose estimation~\cite{zhou2017towards, yang20183d}, and multi-view images based 3D pose estimation~\cite{rhodin2018learning}. 
Human3.6M dataset~\cite{h36m_pami} plays an important role in the passive 3D human pose estimation methods.
It is collected in a highly constrained environment with limited subjects, and background variations.
The innovation of these methods mainly focuses on new techniques that explicitly address the generalization issues when using this dataset.
%
As shown in Figure~\ref{intro:3D skeletion}, the current methods are still problematic when tested against the in-the-wild images.
%
%

To solve the problem, we can improve the generalization ability with well-annotated in-the-wild 3D pose data.
Rogez~\etal~\cite{rogez2016mocap} propose a method to solve the limitations of the laboratory 3D datasets by artificially composing different images to generate a synthetic one based on the 3D Motion Capture (MoCap) data. 
However, the details and variety level of these synthetic images are limited compared with the in-the-wild images.
In this paper, we introduce a principled method to generate high quality 3D labels of the in-the-wild images.
Inspired by~\cite{rhodin2018learning}, to solve the depth ambiguity problem in 3D human pose estimation, we devise a stereo inspired 3D label generator utilizing the 2D poses from multi-view to generate a high quality 3D human pose.
%
%
%
We also propose a geometric search scheme to further refine the predicted 3D human pose. 
%
%
Given any image with the 2D ground truth, the proposed 3D label generator can produce its high quality counterpart.
%
%

%
To this end, based on the 3D label generator, we collect more than 400,000 in-the-wild images with high quality 3D labels from the wildly used 2D pose datasets~\cite{andriluka14cvpr,johnsonclustered,wu2017ai}.
With the proposed in-the-wild 3D pose dataset, we train a high performance baseline network which achieves favorable results against the state-of-the-art methods
, both quantitatively and qualitatively.
Furthermore, we introduce a method that utilizes the predicted 3D human pose on the task of action classification to evaluate the generalization ability quantitatively.
%
%
%

Our contributions can be summarized as follows:
\begin{compactitem}
   \item We propose a novel stereo inspired neural network to generate high quality 3D pose labels for in-the-wild images. We also devise a geometric searching scheme to further refine the 3D joints.
   \item We build a large-scale dataset with 400,000 in-the-wild images and the corresponding high quality 3D pose labels.
   %
   \item We train a baseline network with the proposed dataset that performs favorably against the state-of-the-art approaches, both quantitatively and qualitatively. 
   Experimental results demonstrate that the proposed dataset can significantly boost the generalization performance on the realistic scenes.
\end{compactitem}

\begin{figure*}[!t]
\footnotesize
\centering
\includegraphics[width=\linewidth]{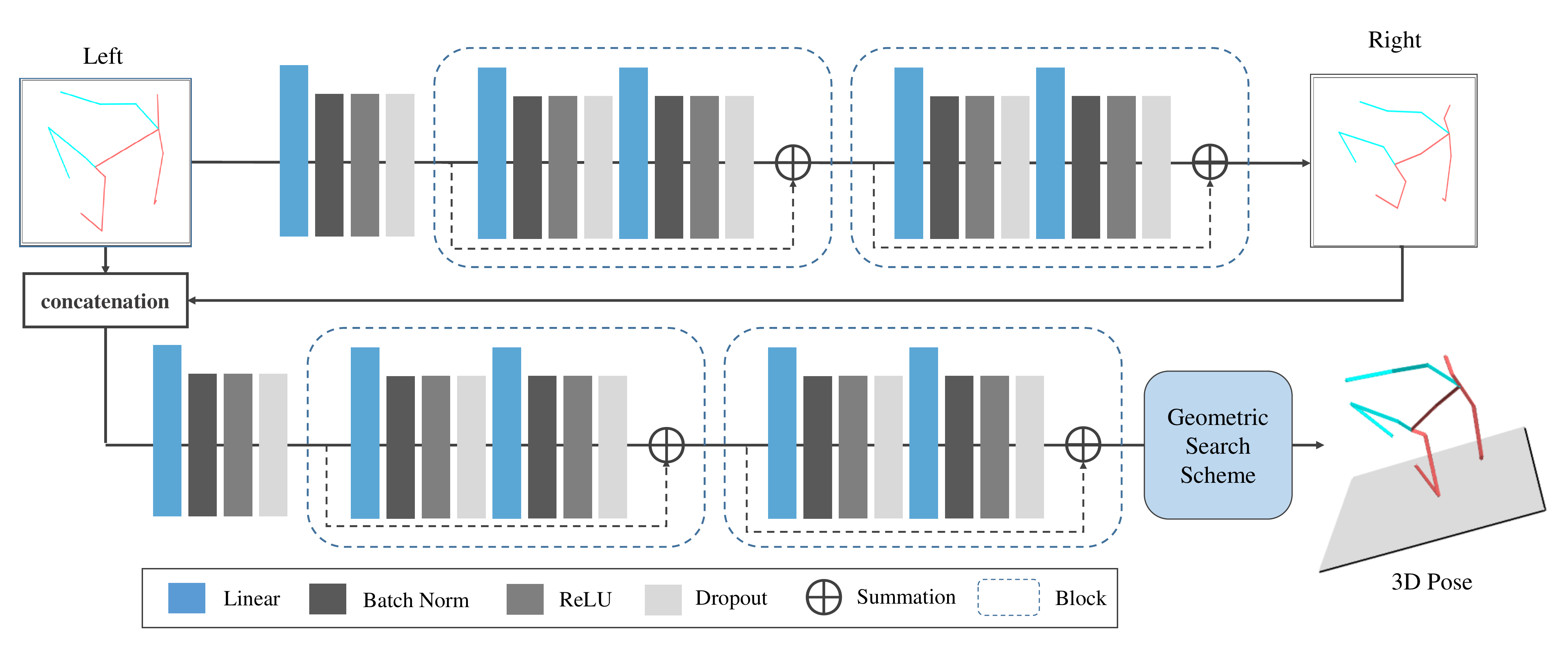}
\vspace{-2mm}
\caption{Architecture of the 3D label generator. 
The generator consists of stereoscopic view synthesis subnetwork, 3D pose reconstruction subnetwork, and a geometric search scheme.
Given the 2D pose from the left viewpoint, stereoscopic view synthetic subnetwork aims to generate the 2D pose from the right viewpoint.
3D pose reconstruction subnetwork utilizes the multi-view 2D poses to estimate a coarse 3D human pose.
Geometric search scheme is applied to further refine the predicted 3D human pose.}
\vspace{-2mm}
\label{intro:pipline}
\end{figure*}

\section{Related Work}
%
%
\paragraph{Synthetic Images and Additional Annotations.}
Most of the existing 3D pose datasets such as Human3.6M~\cite{h36m_pami}, and HumanEva~\cite{sigal2010humaneva} are collected in indoor scenes and cannot cover various activities.
To solve it, several methods attempt to use the graphics methods~\cite{chen2016synthesizing, mehta2017monocular, pishchulin2012articulated, varol2017learning} to enrich the training samples.
Rogez and Schmid~\cite{rogez2016mocap} introduce a computer graphics engine that artificially composes different images to generate synthetic poses based on the 3D Motion Capture (MoCap) data.
Recently, to alleviate the need for the accurate 3D labels of in-the-wild images, Pavlakos~\etal~\cite{pavlakos2018ordinal} and Shi~\etal~\cite{shi2018fbi} provide in-the-wild images with additional annotations which contain the forward or backward information of each bone. 
However, the aforementioned methods either have limited details and variety level of the synthetic images or require a large number of manual annotations.
%
Different from them, we propose a network to automatically generate a large-scale in-the-wild 3D human pose dataset.
%
%
%
\paragraph{2D-to-3D Pose Estimation.}\vspace{-2mm}
Several methods tackle 3D pose understanding from 2D pose~\cite{ akhter2015pose,fan2014pose,lee1985determination, ramakrishna2012reconstructing, taylor2000reconstruction, wang2014robust, yasin2016dual, zhou20153d}.
Martinez~\etal~\cite{martinez2017simple} propose a simple multilayer perceptron to regress the locations of the 3D joints.
Fang~\etal~\cite{fang2017learning} introduce a model to encode the mapping function of human pose from 2D to 3D by explicitly encoding human body configuration with pose grammar. 
%
Despite the consideration of the domain knowledge of the human body, models trained with 2D/3D key-points from Human3.6M containing only fifteen activities cannot perform well to various actions. 
Training with 2D/3D pairs including more than 2,500 activities generated by the unity toolbox~\footnote{\url{https://unity3d.com}}, we devise a network can map an in-the-wild 2D pose to its high quality 3D counterpart. 
\paragraph{Monocular Image Based 3D Pose Estimation.}\vspace{-2mm}
Recently, several methods have been proposed to estimate the 3D pose on the monocular image~\cite{ionescu2014iterated,li20143d, mehta2017monocular, pavlakos2017coarse, sun2017compositional, tekin2017learning, tome2017lifting, zhou2016deep}.
To improve the generalization on realistic scenes, some attempt to estimate 3D human pose in a semi-supervised way.
Zhou~\etal~\cite{zhou2017towards} employ a weakly-supervised transfer learning method with a 3D geometric loss.
Yang~\etal~\cite{yang20183d} propose an adversarial learning framework, which distills the 3D human pose structures learned from the fully annotated dataset to in-the-wild images with only 2D pose annotations.
%
%
However, their mechanisms are to transfer the domain knowledge of the constrained dataset to the in-the-wild images without adding new subjects or more activities.
%
%
Trained with our proposed in-the-wild 3D dataset, the network performs better on the in-the-wild images.
\paragraph{Multi-View Images Based 3D Pose Estimation.}\vspace{-2mm}
Some methods attempt to estimate the 3D human pose from multiple views of different cameras~\cite{amin2013multi,hofmann2012multi,holte2012human,rhodin2018learning}.
Amin~\etal~\cite{amin2013multi} propose the evidence across multiple viewpoints to allow for robust 3D pose
estimation.
Rhodin~\etal~\cite{rhodin2018learning} propose to predict the same 3D pose in all views with only a small number of labeled images.
However, compared with generating 2D joints from different cameras, obtaining the images from multi-views are more difficult. 
Based on these methods, we devise a network with the simple 2D joints from the multi-view to generate the high quality 3D human poses.
\section{Methodology}
\label{sec:meth}
In this section, we first introduce the principles of the network design.
As shown in Figure~\ref{intro:pipline}, we present a novel 3D label generator, which consists of stereoscopic view synthesis subnetwork, 3D pose reconstruction subnetwork, and a geometric search scheme.
In addition, we propose a large-scale in-the-wild 3D pose dataset, and its 3D pose labels are provided by the 3D pose generator.
Furthermore, we adopt a baseline network to evaluate the proposed in-the-wild 3D pose dataset.



%
\subsection{Principles of Network Design}
\label{meth:ple}
2D-to-3D human pose estimation inherently accompanies with the depth ambiguity since the mapping function from 2D to 3D is not unique.
Amin~\etal~\cite{amin2013multi} propose the evidence across multiple viewpoints to allow for robust 3D pose estimation.
Recently, Luo~\etal~\cite{luo2018single} confirm that utilizing images from two different cameras can achieve excellent performance in the stereo matching area.
Inspired by~\cite{luo2018single}, we devise the stereo inspired neural network utilizing the 2D key-points from two different viewpoints to alleviate the depth ambiguity of predicting 3D poses.
Different from the previous methods~\cite{rhodin2018learning, amin2013multi} using multi-view images as inputs to estimate the 3D pose, our method is relatively easier to obtain the training data, since the unity toolbox can generate a large number of 2D/3D pairs automatically.

In addition, most existing 2D-to-3D methods~\cite{hossain2018exploiting,fang2017learning} mainly focus on the domain-knowledge of the human body or the architecture of the network while ignoring that a reasonable predicted 3D human pose can be re-projected to its 2D input with zero-pixel error.
Based on this principle, we devise a geometric search scheme to further refine the predicted coarse 3D human pose.
%
\subsection{Stereoscopic View Synthesis Subnetwork}
Stereoscopic view synthesis subnetwork is proposed to synthesize the 2D pose from the right viewpoint .
Given an image with 2D key-points from the left viewpoint, we generate the 2D key-points from the right viewpoint.
As shown in Figure~\ref{intro:pipline}, we input the left-view 2D pose $(u_{L}, v_{L})$ to regress the location of the right-view 2D pose {$(u_{R}, v_{R})$}.
However, the challenge is how to obtain the ground truth of the 2D pose from the right viewpoint .

We employ a large bunch of 3D key-points and their corresponding camera intrinsic matrix from the realistic 3D pose dataset (\ie Human3.6M~\cite{h36m_pami}) and the synthetic data generated by the unity toolbox to train the subnetwork.
To obtain the ground truth of the right-view 2D pose, we move 3D joints to the right direction slightly along the $X$ axis in the camera coordinate system while keep $Y$ and $Z$ unchanged. We then re-map them into the 2D key-points based on the camera calibration by the following equation:
\begin{align}
\label{equ:map2d}
s\begin{bmatrix}u_{R}
\\ v_{R}
\\ 1
\end{bmatrix} &= \begin{bmatrix}
 \alpha_{x}& 0 &u_{0}  \\ 
 0& \alpha_{y}& v_{0}  \\ 
 0& 0&  1  
\end{bmatrix}\begin{bmatrix}
x_{c} + \Delta{x}\\y_{c} 
\\z_{c}  
\end{bmatrix}=M_{c}P_{c},
\end{align}
where $s$ denotes the scale factor, $(u_{R},v_{R})$ represents the right-view 2D pose, $\alpha_{x}$ and $\alpha_{y}$ are the scale factors, $(u_{0}, v_{0})$ denotes the origin coordinate of the RGB image. $P_{c}=(x_{c},y_{c},z_{c})$ is the location of 3D key-points in the camera coordinate system. $M_{c}$ represents the camera intrinsic matrix. $\Delta_{x} = 500\mathrm{mm}$ denotes the shift distance. 

%
%
The subnetwork contains the linear-ReLU layers, residual connections, batch normalization layers, and dropout layers with max-norm constraint.
\subsection{3D Pose Reconstruction Subnetwork}
As shown in Figure~\ref{intro:pipline}, 3D pose reconstruction subnetwork directly regresses the location of 3D key-points based on the input left-view 2D pose and synthesized right-view 2D pose.
It shares the same architecture as the stereoscopic view synthesis subnetwork and takes the left-view and synthetic right-view 2D poses as the inputs.
More precisely, when inputting the multi-view 2D poses, we combine them by the concatenation operation.
After the operation of a fully connected network structure, we obtain a coarse 3D human pose.
The 3D pose reconstruction can be represented by the following equation:
\begin{equation}
Q_{r}=f_{r}((u_{L}, v_{L}), (u_{R}, v_{R})),
\end{equation}
where $Q_{r} = (x_{r},y_{r},z_{r})\in\mathbb{R}^{3\times{N}}$ denotes the predicted 3D human pose by the 3D pose reconstruction subnetwork, , $N$ denotes the joint number. 
\subsection{Geometric Search Scheme}

\begin{figure}[t]
\footnotesize
\centering
\includegraphics[width=\linewidth]{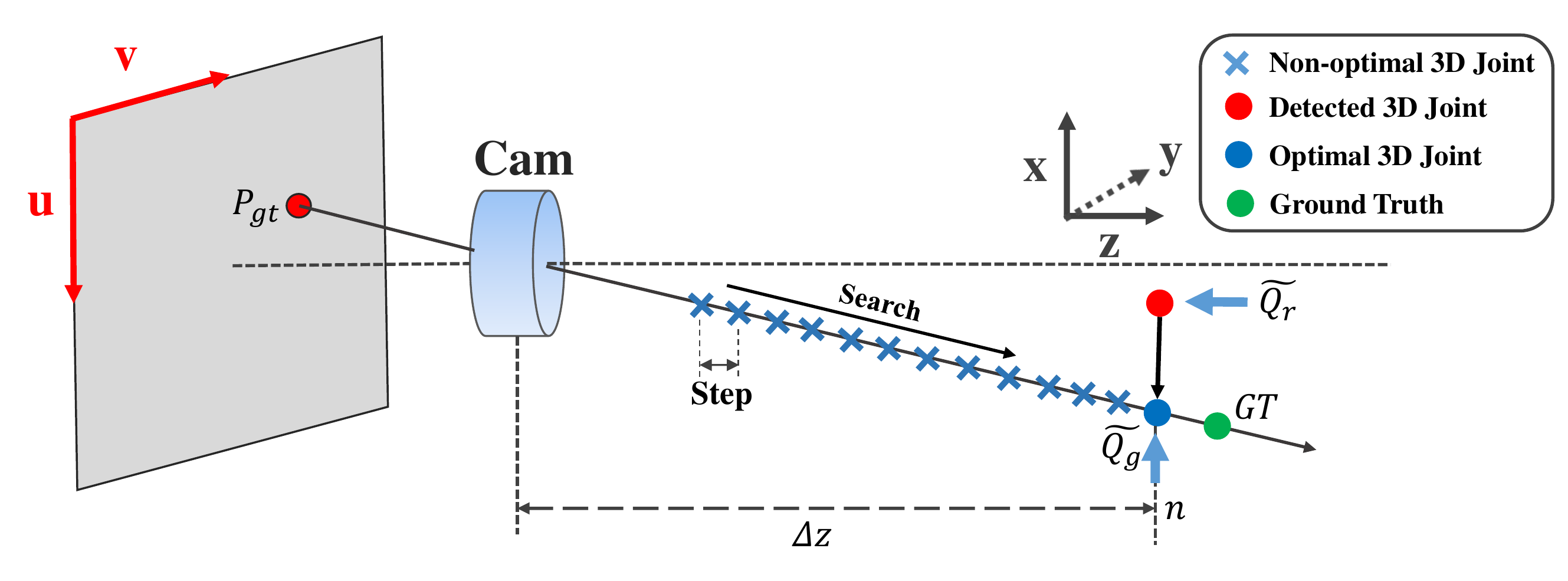}
\vspace{-2mm}
\caption{Geometric search scheme. $\widetilde{Q_{r}}=(\widetilde{x_{r}},\widetilde{y_{r}},\widetilde{z_{r}})$ denotes the predicted 3D human pose by the 3D pose reconstruction subnetwork with hypothetical depth to the camera. 
$\widetilde{Q_{g}}=(\widetilde{x_{g}},\widetilde{y_{g}},\widetilde{z_{g}})$ represents the 3D human pose with the absolute depth to the camera.}
\vspace{-6mm}
\label{meth:gsm}
\end{figure}
The geometric search scheme aims to further refine the coarse 3D human pose.
It ensures the refined 3D human pose can be projected to the input 2D joints via the camera intrinsic matrix with zero-pixel error.
Mathematically, our geometric search scheme can be represented by the following equation:
\begin{equation}
Q_{g}=f_{geo}(P_{gt},Q_{r}),
\end{equation}
where $P_{gt} = (u_{gt},v_{gt})\in\mathbb{R}^{2\times{N}}$ represents ground truth of the real 2D pose, $Q_{g} = (x_{g},y_{g},z_{g})\in\mathbb{R}^{3\times{N}}$ is the final output of the 3D label generator.
Actually, the 3D pose reconstruction subnetwork outputs the 3D human pose aligned to the root joint (pelvis).
What our model predicts in our case is 3D key-points with relative depth (relative to hip). Therefore, the projection is not possible because it requires absolution depth.
According to the camera calibration principle, we propose the heuristic projection to constrain the consistence between the input 3D pose and projected 2D pose.
Figure~\ref{meth:gsm} shows the procedure of the geometric search scheme.
Based on the $z_{r}$ and $P_{gt}$, we initialize $\Delta{z}=0$ ($\Delta{z}$ represents the hypothetical depth to the camera).
Then We can infer the $\widetilde{x}_{r}$ and $\widetilde{y}_{r}$ according to the camera intrinsic matrix and $\Delta{z}$, and the process of searching the optimal 3D joint can be described as follows:
\begin{align}
\nonumber  \widetilde{x}_{r} &= (u_{gt} - c_{x})(z_{r} + \Delta{z}) / f_{x} \\
           \widetilde{y}_{r} &= (v_{gt} - c_{y})(z_{r} + \Delta{z}) / f_{y}
\end{align}
, where $f_{x}$ and $f_{y}$ denote focal length of the camera and $c_{x}$ and $c_{y}$ represent the optical axis points of the camera.
By increasing $\Delta{z}$ with $step = 1\mathrm{mm}$, until we obtain the optimal value that can satisfy the following loss function,
\begin{equation}
    L_{geo}=\mathop{\arg\min}_{\Delta{z}}\left\|((\widetilde{x}_{r}-x_{r})^{2}+(\widetilde{y}_{r}-y_{r})^{2})\right\|_{2}^{2}
\end{equation}
In this way, we can obtain a reasonable $\Delta{z}$, the location of the 3D pose $\widetilde{Q}_{g}$ in the 3D space, and the final output of the 3D label generator $Q_{g}$. 
\subsection{A Large-Scale in-the-Wild 3D Pose Dataset}
The proposed 3D label generator can map a 2D human pose to its high quality 3D counterpart.
The existing datasets for 2D human pose estimation such as Leeds Sports Pose dataset (LSP)~\cite{johnsonclustered}, MPII human pose dataset (MPII)~\cite{andriluka14cvpr} and Ai Challenger dataset for 2D human pose estimation (Ai-Challenger)~\cite{wu2017ai} can be used to extract the high quality 3D labels by the 3D label generator.
We crop the single person from the 2D datasets and set each human body to be in the center of the image.
Then we resize these images into $256\times256$.
Given the well-annotated 2D key-points, the 3D label generator can output the high quality 3D labels of the in-the-wild images.
Finally, we collect a large-scale in-the-wild dataset containing more than 400,000 images (320,000 training images and the rest for testing) with high quality 3D labels.
Experimental results demonstrate that the proposed in-the-wild 3D pose dataset can improve the performance of 3D human pose estimation quantitatively and qualitatively.

\begin{table*}[t]
\begin{center}
\caption{Quantitative evaluations on the Human3.6M~\cite{h36m_pami} under Protocol\#1 (no rigid alignment or similarity transform is applied in post-processing). GT indicates that the network was trained on ground truth 2D pose. GS denotes the geometric search scheme. Unity denotes the model trained with the additional 2D/3D key-points generated by the unity toolbox. The bold-faced numbers represent the best results.}\vspace{2mm}
\label{exp:pro1}
\resizebox{\textwidth}{!}{%
\begin{tabular}{lcccccccccccccccc}
\toprule
\textbf{Protocol\#1} ($\downarrow$)        & Direct. & Discuss & Eating & Greet & Phone & Photo & Pose & Purch. & Sitting & SittingD. & Smoke & Wait & WalkD & Walk & WalkT. & Average \\ \midrule
LinKDE~\cite{h36m_pami}                    & 132.7   & 183.6   & 132.3  & 164.4 & 162.1 & 205.9 & 150.6 & 171.3 & 151.6 & 243.0 & 162.1 & 170.7 & 177.1 & 96.6 & 127.9 & 162.1 \\
Tekin~\etal~\cite{tekin2016direct}         & 102.4   & 147.2   & 88.8   & 125.3 & 118.0 & 182.7 & 112.4 & 129.2 & 138.9 & 224.9 & 118.4 & 138.8 & 126.3 & 55.1 & 65.8  & 125.0 \\
Zhou~\etal~\cite{zhou2016sparseness}       & 87.4    & 109.3   & 87.1   & 103.2 & 116.2 & 143.3 & 106.9 & 99.8  & 124.5 & 199.2 & 107.4 & 118.1 & 114.2 & 79.4 & 97.7  & 113.0 \\
Park~\etal~\cite{park20163d}               & 100.3   & 116.2   & 90.0   & 116.5 & 115.3 & 149.5 & 117.6 & 106.9 & 137.2 & 190.8 & 105.8 & 125.1 & 131.9 & 62.6 & 96.2  & 117.3 \\  
Nie~\etal~\cite{nie2017monocular}          & 90.1    & 88.2    & 85.7   & 95.6  & 103.9 & 103.0 & 92.4  & 90.4  & 117.9 & 136.4 & 98.5  & 94.4  & 90.6  & 86.0 & 89.5  & 97.5  \\  
Metha~\etal~\cite{mehta2016monocular}      & 57.5    & 68.6    & 59.6   & 67.3  & 78.1  & 82.4  & 56.9  & 69.1  & 100.0 & 117.5 & 69.4  & 68.0  & 76.5  & 55.2 & 61.4  & 72.9  \\  
Metha~\etal~\cite{mehta2017vnect}          & 62.6    & 78.1    & 63.4   & 72.5  & 88.3  & 93.8  & 63.1  & 74.8  & 106.6 & 138.7 & 78.8  & 73.9  & 82.0  & 55.8 & 59.6  & 80.5  \\  
Lin~\etal~\cite{lin2017recurrent}          & 58.0    & 68.2    & 63.3   & 65.8  & 75.3  & 93.1  & 61.2  & 65.7  & 98.7  & 127.7 & 70.4  & 68.2  & 72.9  & 50.6 & 57.7  & 73.1  \\  
Tome~\etal~\cite{tome2017lifting}          & 65.0    & 73.5    & 76.8   & 86.4  & 86.3  & 110.7 & 68.9  & 74.8  & 110.2 & 173.9 & 84.9  & 85.8  & 86.3  & 71.4 & 73.1  & 88.4  \\  
Tekin~\etal~\cite{tekin2017learning}       & 54.2    & 61.4    & 60.2   & 61.2  & 79.4  & 78.3  & 63.1  & 81.6  & 70.1  & 107.3 & 69.3  & 70.3  & 74.3  & 51.8 & 63.2  & 69.7  \\  
Pavlakos~\etal~\cite{pavlakos2017coarse}   & 67.4    & 71.9    & 66.7   & 69.1  & 72.0  & 77.0  & 65.0  & 68.3  & 83.7  & 96.5  & 71.7  & 65.8  & 74.9  & 59.1 & 63.2  & 71.9  \\  
Martinez~\etal~\cite{martinez2017simple}   & 51.8    & 56.2    & 58.1   & 59.0  & 69.5  & 78.4  & 55.2  & 58.1  & 74.0  & 94.6  & 62.3  & 59.1  & 65.1  & 49.5 & 52.4  & 62.9  \\  
Fang~\etal~\cite{fang2017learning}         & 50.1    & 54.3    & 57.0   & 57.1  & 66.6  & 73.3  & 53.4  & 55.7  & 72.8  & 88.6  & 60.3  & 57.7  & 62.7  & 47.5 & 50.6  & 60.4  \\  
Sun~\etal~\cite{sun2017compositional}      & 52.8    & 54.8    & 54.2   & 54.3  & 61.8  & 67.2  & 53.1  & 53.6  & 71.7  & 86.7  & 61.5  & 53.4  & 61.6  & 47.1 & 53.4  & 59.1  \\  
Yang~\etal~\cite{yang20183d}               & 51.5    & 58.9    & 50.4   & 57.0  & 62.1  & 65.4  & 49.8  & 52.7  & 69.2  & 85.2  & 57.4  & 58.4  & 43.6  & 60.1 & 47.7  & 58.6  \\ \midrule
Martinez~\etal~\cite{martinez2017simple} (GT) w/o GS         & 37.7    & 44.4    & 40.3   & 42.1  & 48.2  & 54.9  & 44.4  & 42.1  & 54.6  & 58.0  & 45.1  & 46.4  & 47.6  & 36.4 & 40.4  & 45.5  \\
Martinez~\etal~\cite{martinez2017simple} (GT) w/ GS          & 33.1    & 39.8    & 34.5   & 37.5  & 39.5  &\textbf{45.7}  & 40.4  & 31.7  & 44.9  & 49.2  & 37.8  & 39.2  & 39.8  & \textbf{30.3} & 33.8  & 38.5  \\
Ours (GT) w/o GS                           & 35.6    & 41.3    & 39.4   & 40.0  & 44.2  & 51.7  & 39.8  & 40.2  & 50.9  & 55.4  & 43.1  & 42.9  & 45.1  & 33.1 & 37.8  & 42.0  \\
Ours (GT) w/ GS & \textbf{32.1} & \textbf{39.2} & \textbf{33.4} & \textbf{36.4} & \textbf{38.9} & 45.9 & \textbf{38.4} & \textbf{31.7} & \textbf{42.5} & \textbf{48.1} & \textbf{37.8} & \textbf{37.9} & \textbf{38.7} & 30.6 & \textbf{32.6} & \textbf{37.6} \\ \midrule
Ours (GT) w/ GS + unity         & 36.5    & 42.7   & 38.2   & 39.6   & 45.3  & 50.8  & 40.2  & 34.8  & 45.0  & 50.3  & 39.4  & 39.9  & 42.5  & 32.2 & 33.8  & 40.8  \\\bottomrule

\end{tabular}}
\vspace{-6mm}
\end{center}
\end{table*}

\subsection{Baseline Network}
We adopt the backbone of Zhou~\etal~\cite{zhou2017towards} as the baseline network to evaluate the in-the-wild 3D pose dataset quantitatively and qualitatively.
This network can be viewed as a two-stage pose estimator.
The first stage is to use the stacked hourglass network~\cite{newell2016stacked} for 2D human pose estimation.
Each stack is in an encoder-decoder structure.
The second stage is a depth regression module.
Given the 2D body joints heat-maps and intermediate features generated from stacked hourglass network, it can predict the depth of each joint.
%
%
%
Since we have a large-scale in-the-wild 3D pose dataset, we discard the weakly-supervised designs employed by Zhou~\etal~\cite{zhou2017towards} and Yang~\etal~\cite{yang20183d}.
We train the network only using the first two stages of the method with Human3.6M and in-the-wild 3D pose dataset.
\section{Experiments}

In this section, we present the experiments and results of the 3D label generator.
Trained with 2D ground truth, our 3D label generator achieves state-of-the-art results on the Human3.6M dataset~\cite{h36m_pami}.
Experimental results denote that the generator can provide high quality labels for the in-the-wild images.
Meanwhile, we investigate the efficacy of the stereoscopic view synthesis subnetwork, 3D pose reconstruction subnetwork, and geometric search scheme respectively.
In addition, we compare with the methods of ~\cite{zhou2017towards,yang20183d} to verify the effectiveness of in-the-wild 3D pose dataset.
To further verify the generalization ability of the model, we attempt to use our predicted 3D poses to the task of classification on the Penn Action dataset~\cite{zhang2013actemes}.
\subsection{Datasets and Evaluation Metrics}
We numerically evaluate the publicly available 3D human pose estimation dataset: Human3.6M~\cite{h36m_pami}.
We also conduct qualitative experiments on in-the-wild images.
\paragraph{3D Pose Datasets.}\vspace{-4mm}
Human3.6M is a large-scale dataset with 2D joint locations and 3D ground truth collected by the MoCap system in the laboratory environment. 
It consists of 3.6 million RGB images of 11 different professional actors performing 15 everyday activities.
Following our baseline method~\cite{zhou2017towards}, we employ data from subjects S1, S5, S6, S7, S8 for training and evaluate on the data from subjects S9 and S11.
We refer the MPJPE that evaluated on the predicted 3D pose after alignment of the root without any rigid alignment transformation as protocol\#1.

MPI-INF-3DPH~\cite{mehta2017monocular} is a recent dataset that includes both indoor and outdoor scenes, which contains 2929 frames from six subjects performing seven actions, to evaluate the generalization ability quantitatively.
We only use the test split of this dataset to demonstrate the generalization ability of the trained model.
\paragraph{2D Pose Datasets.}\vspace{-5mm}
MPII and LSP are the most widely used dataset for 2D human pose estimation.
Ai-Challenger is proposed recently for multi-person 2D pose estimation, which consists of 210,000 images for training, 30,000 images for validation and 60,000 images for testing.
We qualitatively compare the generalization ability on these 2D datasets.
\paragraph{Penn Action Dataset.}\vspace{-5mm}
Penn Action Dataset~\cite{zhang2013actemes} contains 2,326 video sequences of 15 different actions, \eg{pull-up, squat and push-up}, with 1,258 clips for training and 1,068 clips for testing. 
The performance is measured by the mean classification accuracy across the splits~\cite{simonyan2014two}.
\subsection{Implementation Details}

\begin{figure*}[t]
\footnotesize
\centering
\renewcommand{\tabcolsep}{1pt} 
\renewcommand{\arraystretch}{1} 
\begin{center}
\resizebox{\linewidth}{!}{%
\begin{tabular}{cccccccccccc}
 \includegraphics[width=0.08\linewidth]{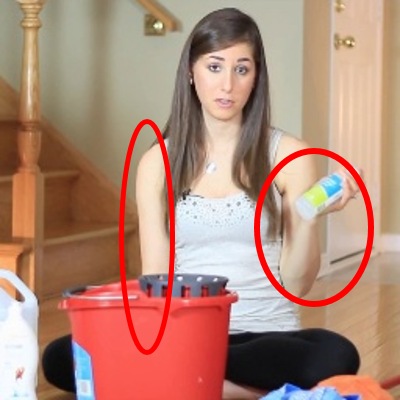}&
 \includegraphics[width=0.08\linewidth]{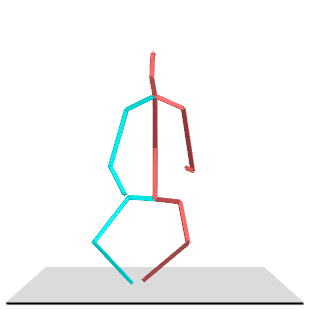}&
 \includegraphics[width=0.08\linewidth]{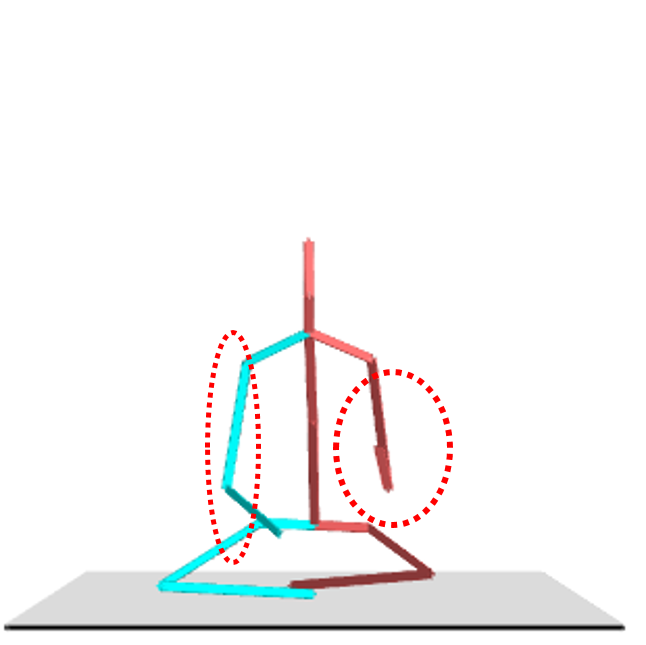}&
 \includegraphics[width=0.08\linewidth]{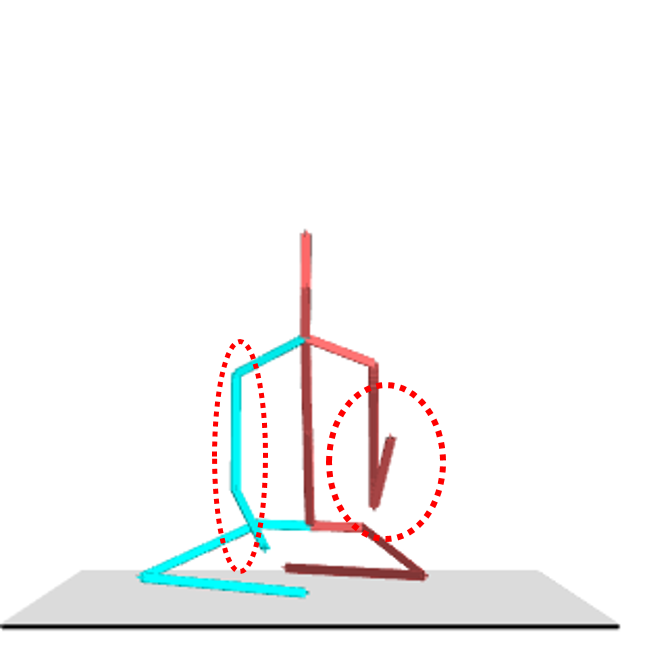}&
 \includegraphics[width=0.08\linewidth]{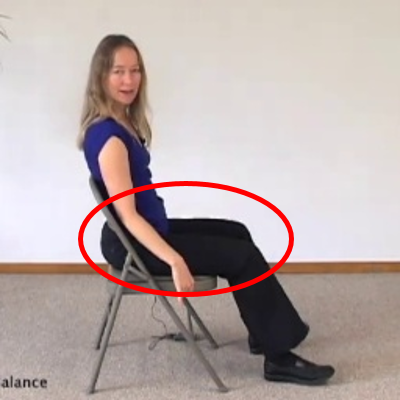}&
 \includegraphics[width=0.08\linewidth]{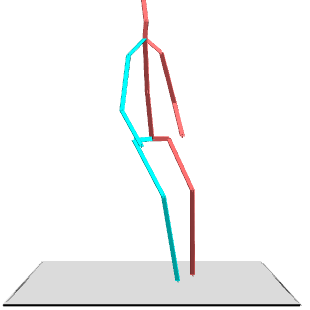}&
 \includegraphics[width=0.08\linewidth]{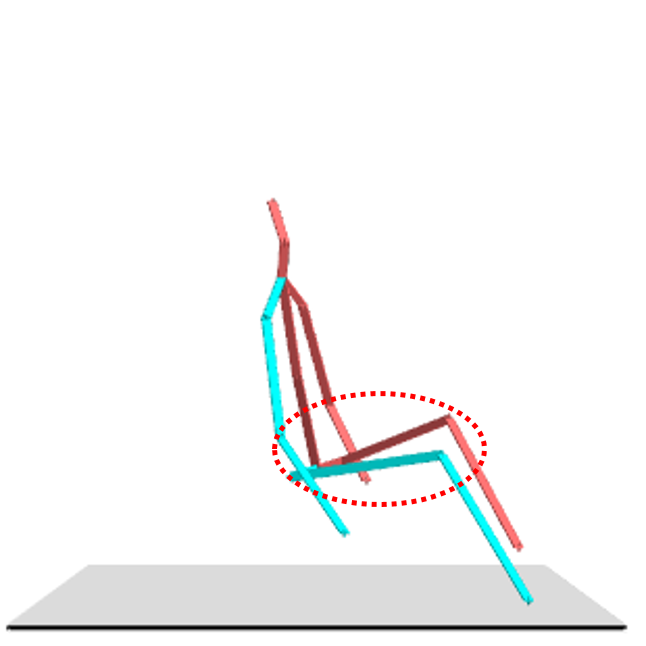}&
 \includegraphics[width=0.08\linewidth]{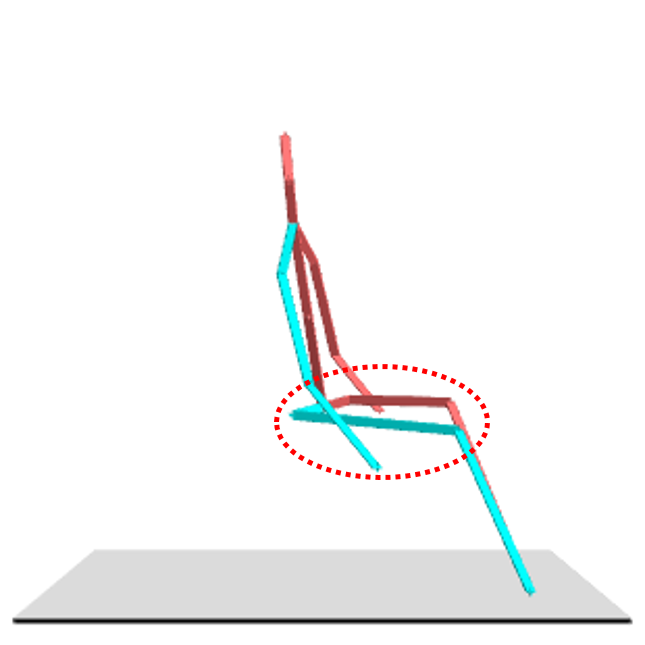}&
 \includegraphics[width=0.08\linewidth]{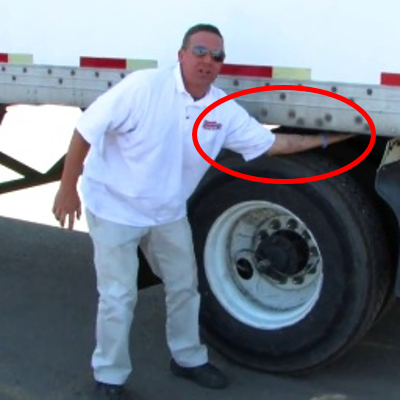}&
 \includegraphics[width=0.08\linewidth]{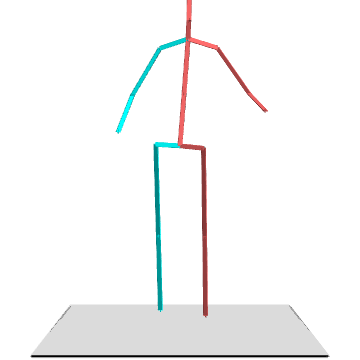}&
 \includegraphics[width=0.08\linewidth]{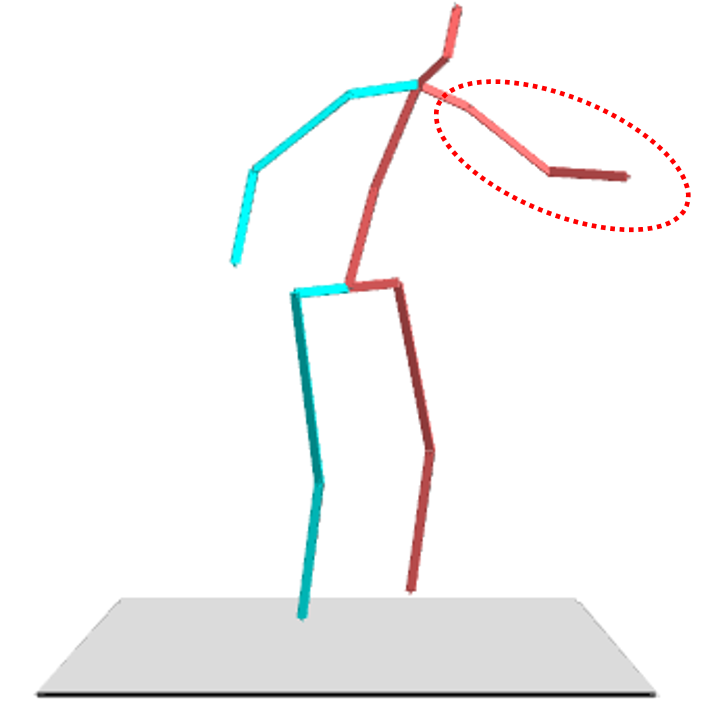}&
 \includegraphics[width=0.08\linewidth]{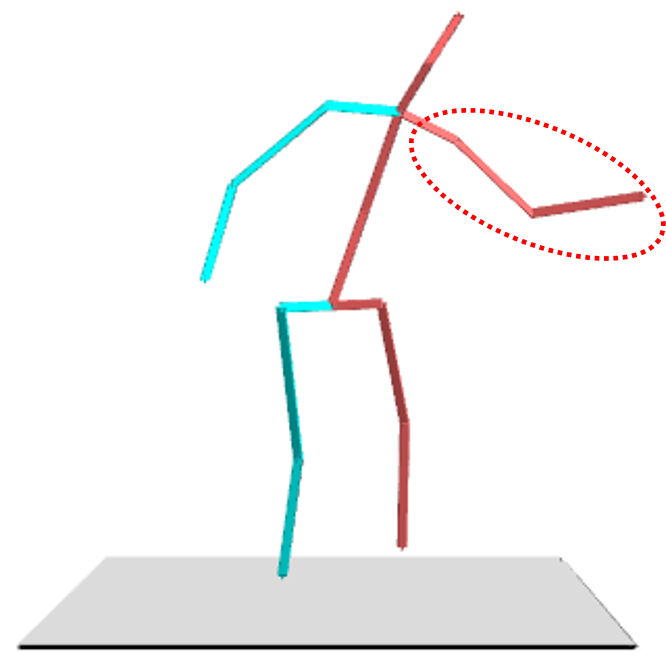}\\

 \includegraphics[width=0.08\linewidth]{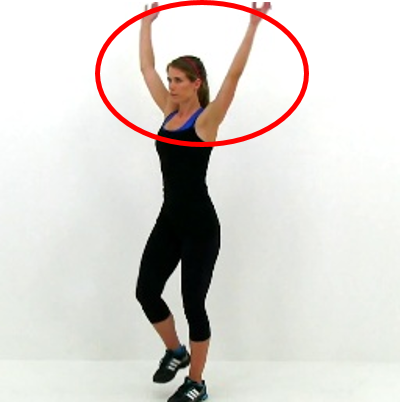}&
 \includegraphics[width=0.08\linewidth]{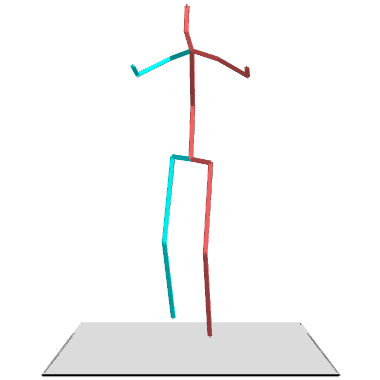}&
 \includegraphics[width=0.08\linewidth]{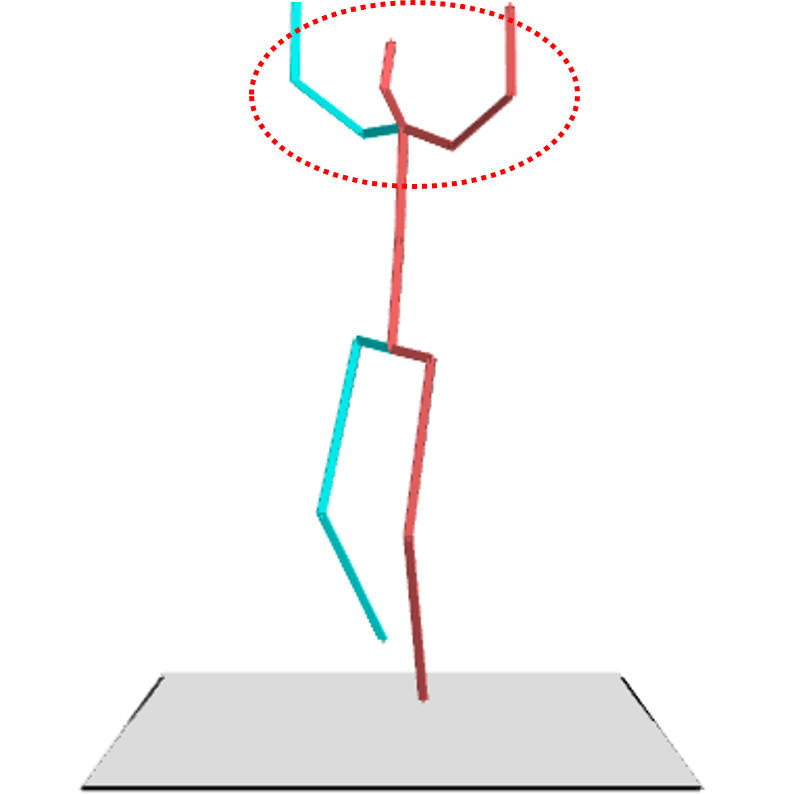}&
 \includegraphics[width=0.08\linewidth]{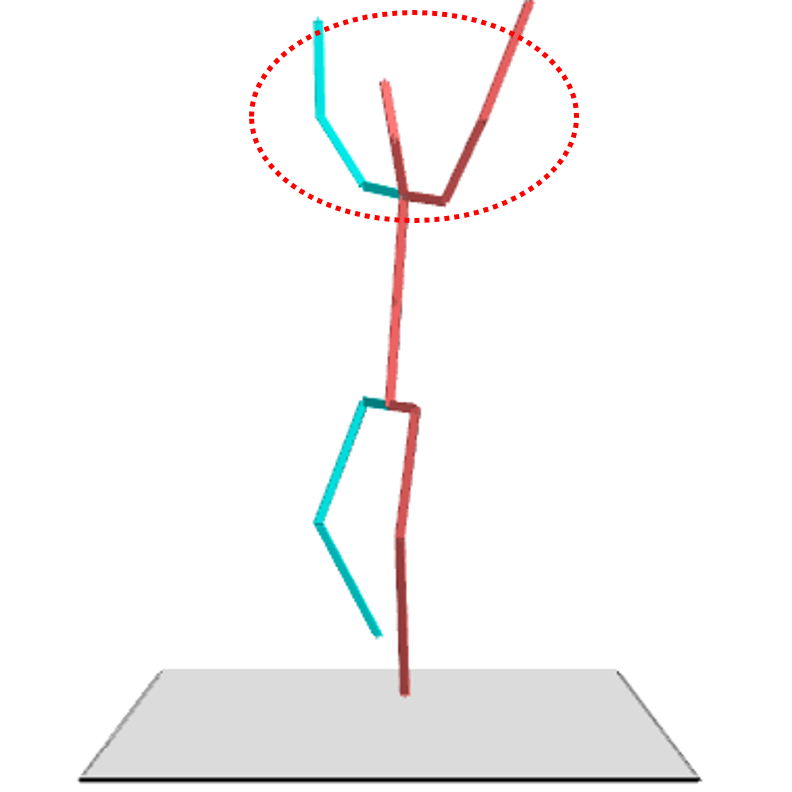}&
 \includegraphics[width=0.08\linewidth]{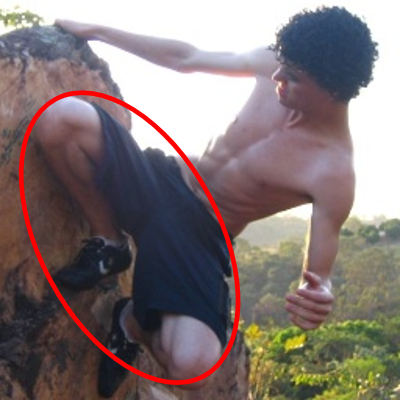}&
 \includegraphics[width=0.08\linewidth]{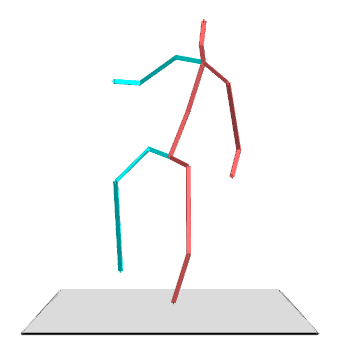}&
 \includegraphics[width=0.08\linewidth]{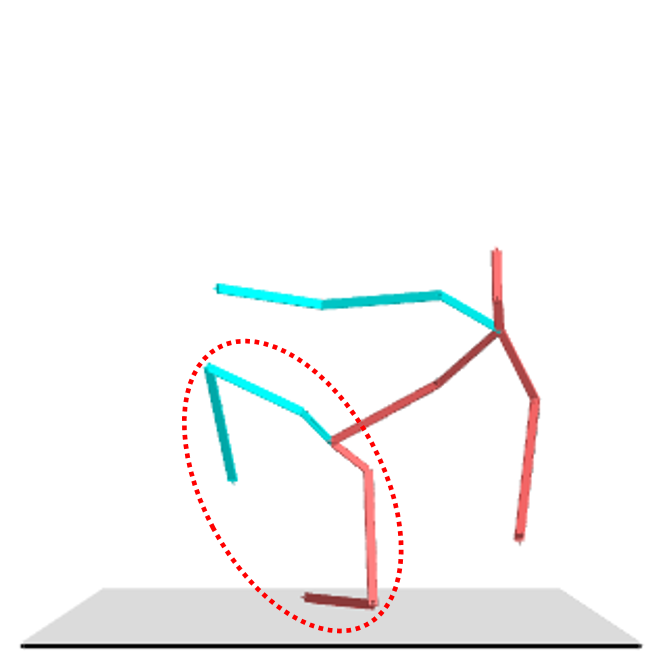}&
 \includegraphics[width=0.08\linewidth]{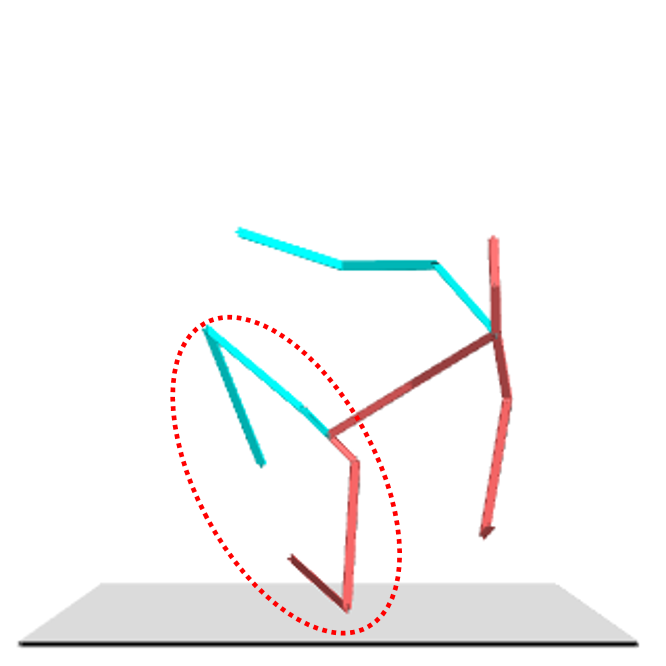}&
 \includegraphics[width=0.08\linewidth]{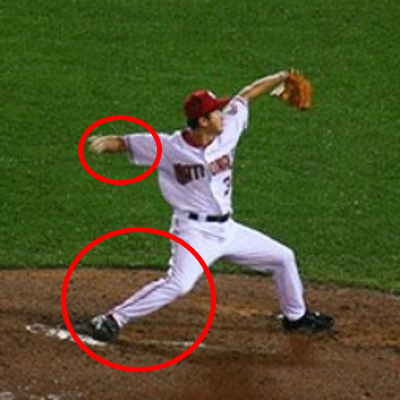}&
 \includegraphics[width=0.08\linewidth]{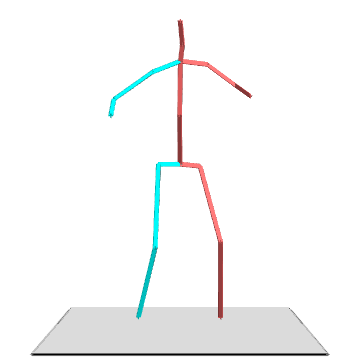}&
 \includegraphics[width=0.08\linewidth]{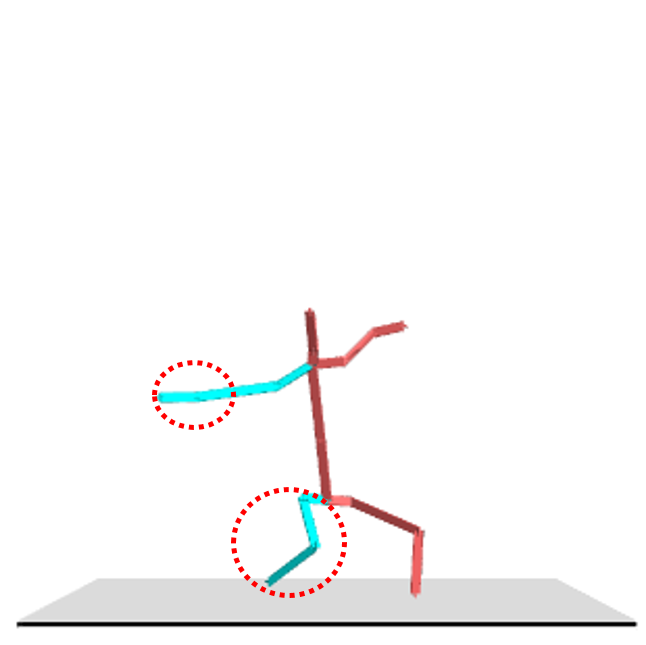}&
 \includegraphics[width=0.08\linewidth]{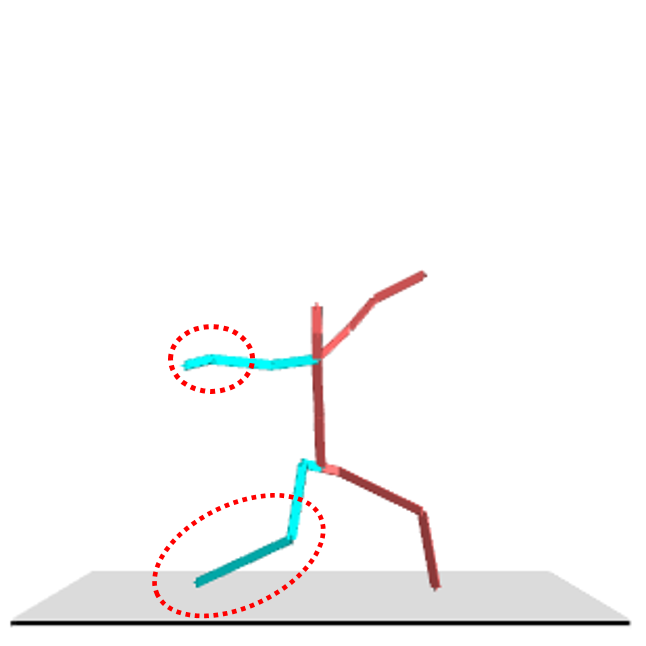}\\

 \includegraphics[width=0.08\linewidth]{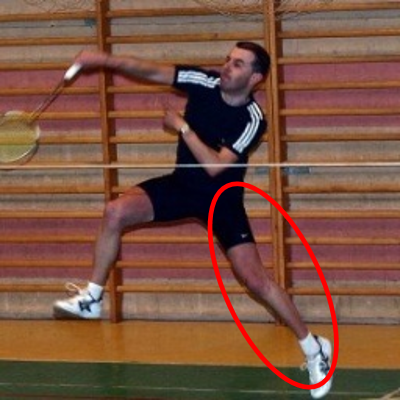}&
 \includegraphics[width=0.08\linewidth]{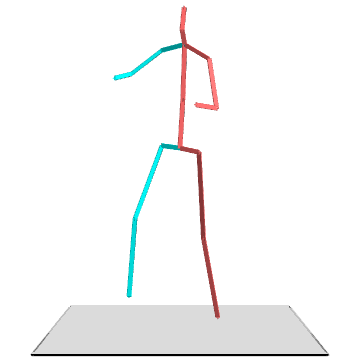}&
 \includegraphics[width=0.08\linewidth]{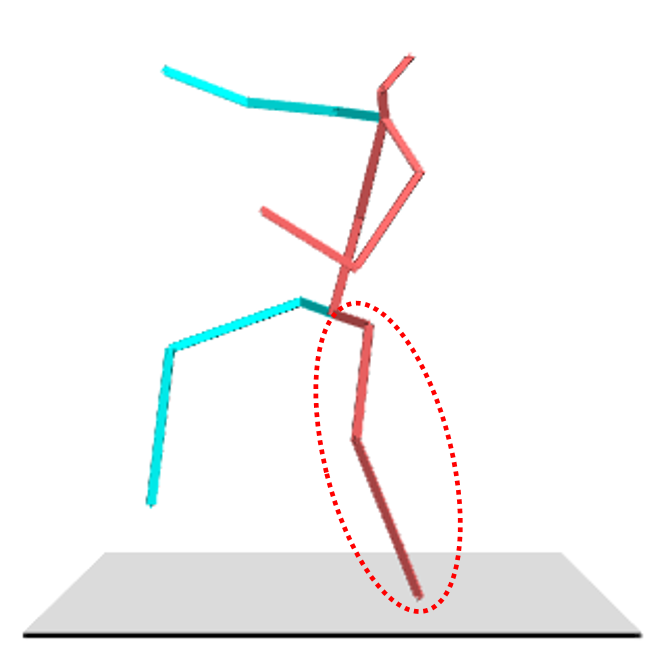}&
 \includegraphics[width=0.08\linewidth]{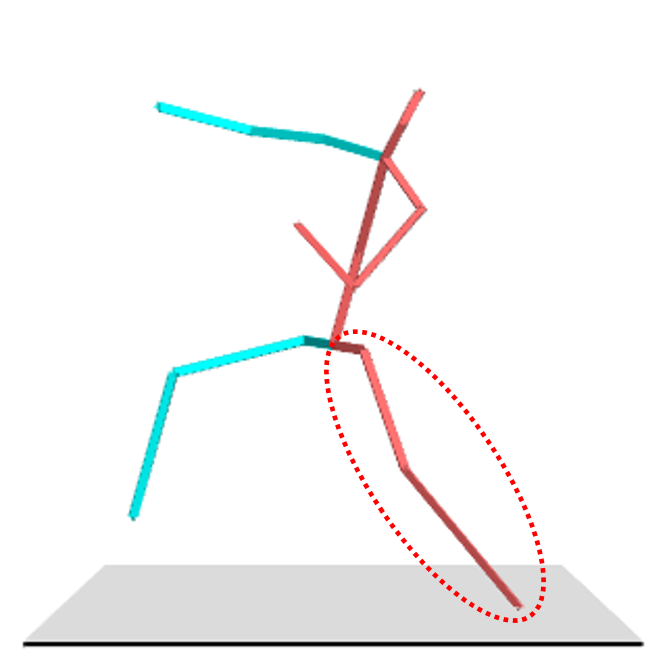}&
 \includegraphics[width=0.08\linewidth]{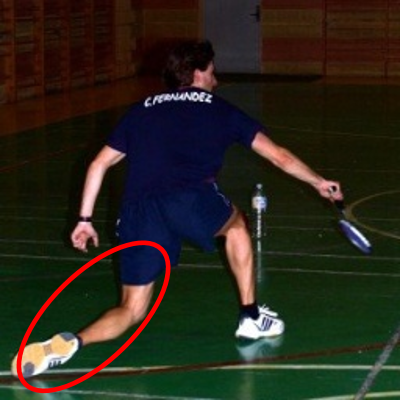}&
 \includegraphics[width=0.08\linewidth]{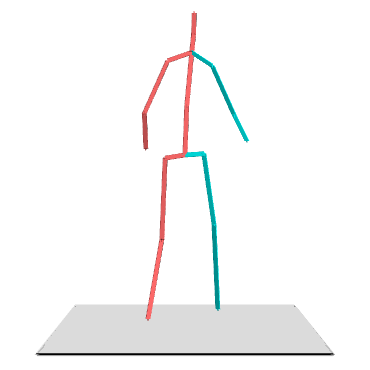}&
 \includegraphics[width=0.08\linewidth]{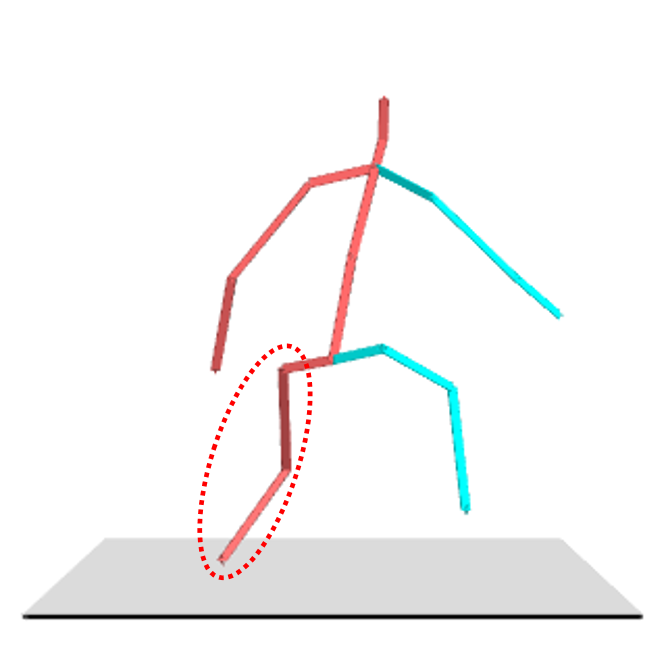}&
 \includegraphics[width=0.08\linewidth]{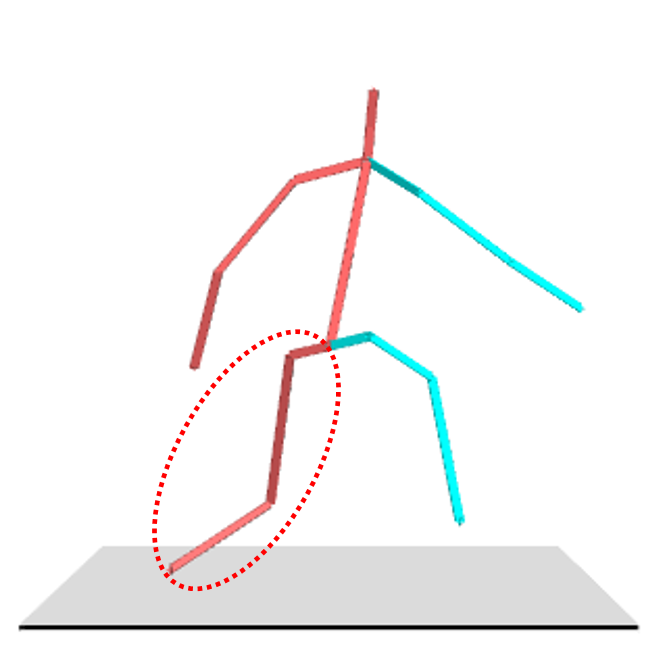}&
 \includegraphics[width=0.08\linewidth]{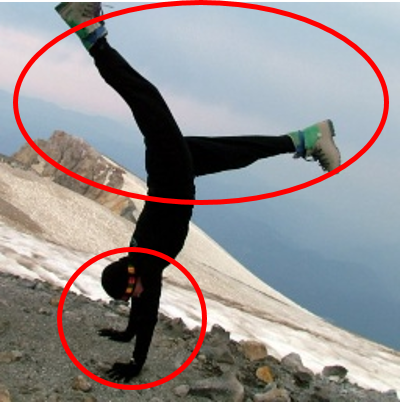}&
 \includegraphics[width=0.08\linewidth]{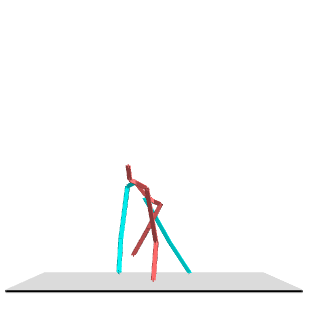}&
 \includegraphics[width=0.08\linewidth]{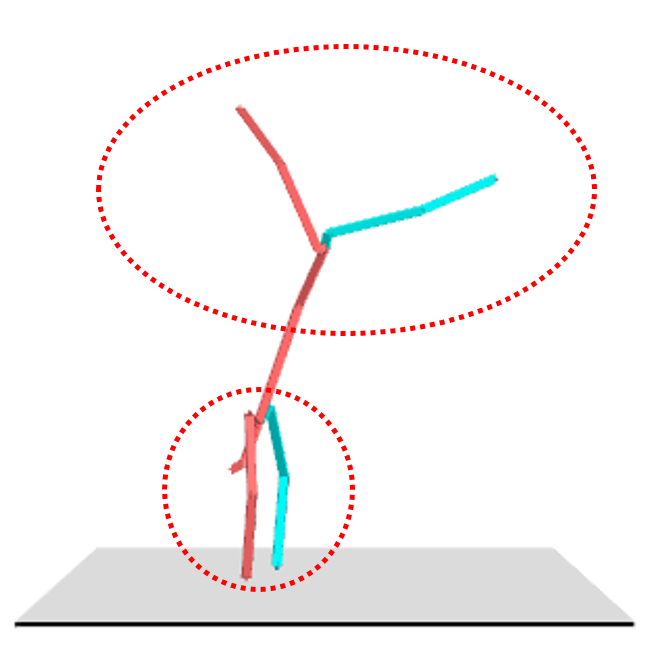}&
 \includegraphics[width=0.08\linewidth]{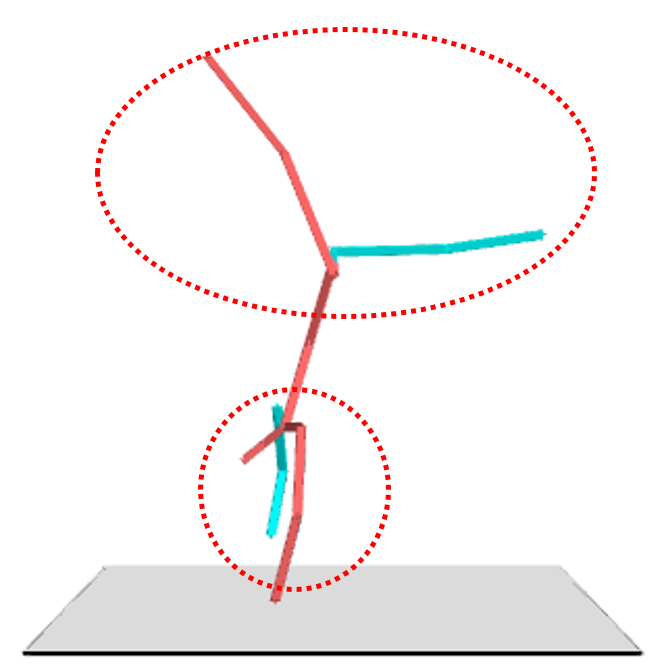}\\

 (a)& (b)& (c)& (d)& (a)& (b)& (c)& (d)& (a)& (b)& (c)& (d)\\
\end{tabular}}
\end{center}
\vspace{-2mm}
\caption{Qualitative evaluations on the in-the-wild images. (a) Original in-the-wild images, (b) Results of Martinez~\etal~\cite{martinez2017simple}, (c) Our results w/o geometric search scheme, (d) Our results w/ geometric search scheme. The proposed 3D label generator outperforms the method of Martinez~\etal~\cite{martinez2017simple}. The proposed geometric search scheme can refine the coarse 3D human pose. (All the predicted 3D poses are demonstrated from the front viewpoint.)}
\vspace{-4mm}
\label{exp:intro}
\end{figure*}

We introduce the implementation details of the 3D label generator and the baseline network based on the RGB images for 3D human pose estimation.
\paragraph{3D Label Generator.}\vspace{-4mm}
We train the proposed 3D label generator using Pytorch~\cite{pytorch} toolbox and Adam~\cite{kinga2015method} solver to optimize the parameters.
We set momentum, momentum2, and weight decay as $0.9$, $0.99$, and $10^{-4}$, respectively.
Kaiming initialization~\cite{he2015delving} is used to initialize the weights of our linear layers.
The network is trained for a total of 200 epoch.
The learning rate is set to be $10^{-3}$ and exponential decay.
We train the stereoscopic view synthesis subnetwork with 4.8 million 2D/3D key-points pairs from the Unity toolbox and Human3.6M.
We set the batch size as 64 and normalize the dataset to $\left [ -1, 1 \right ]$.
Training on a Nvidia TITAN X GPU, the network converges within one day.
When training the 3D pose reconstruction subnetwork, we fix the parameters of the stereoscopic view synthesis subnetwork. 
We adopt the same training scheme when training the 3D pose reconstruction subnetwork.
\paragraph{Baseline Network.}\vspace{-4mm}
Stochastic Gradient Descent (SGD) optimization is used for training.
Each training batch contains both the Human3.6M and in-the-wild images in the ratio of 1:1.
We fine-tune the 2D module based on the checkpoint of Zhou~\etal~\cite{zhou2017towards} with the 2D annotated Human3.6M dataset and the in-the-wild dataset containing MPII, LSP, and Ai-Challenger.
Both data from Human3.6M and the proposed in-the-wild 3D pose dataset are employed for training the two stage baseline network.
We train the network with the loss following Zhou~\etal~\cite{zhou2017towards}.

\subsection{Evaluations on 3D Label Generator}
\begin{figure*}[t]{}
\footnotesize
\centering
\renewcommand{\tabcolsep}{1pt} 
\renewcommand{\arraystretch}{1} 
\begin{center}
\resizebox{\textwidth}{!}{%
\begin{tabular}{cccccccccccc}

 \includegraphics[width=0.08\linewidth]{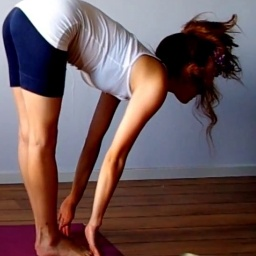}&
 \includegraphics[width=0.08\linewidth]{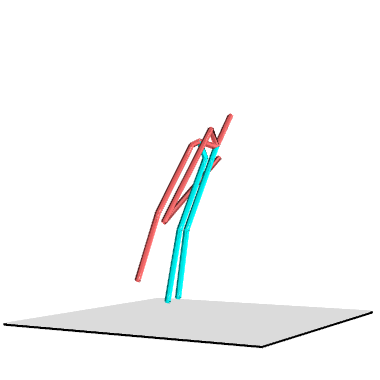}&
 \includegraphics[width=0.08\linewidth]{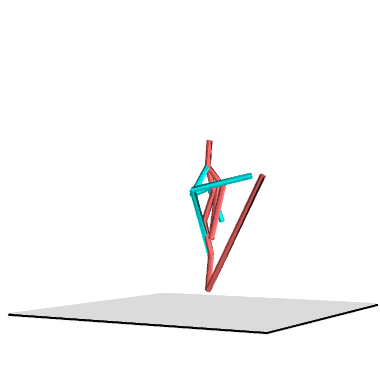}&
 \includegraphics[width=0.08\linewidth]{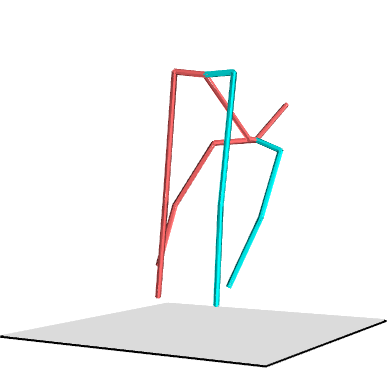}&
 \includegraphics[width=0.08\linewidth]{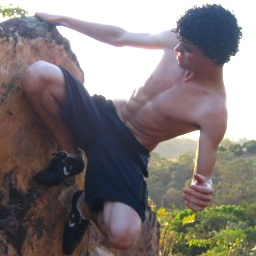}&
 \includegraphics[width=0.08\linewidth]{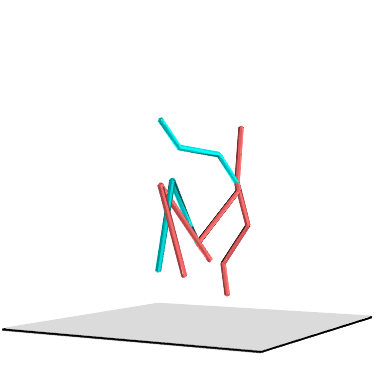}&
 \includegraphics[width=0.08\linewidth]{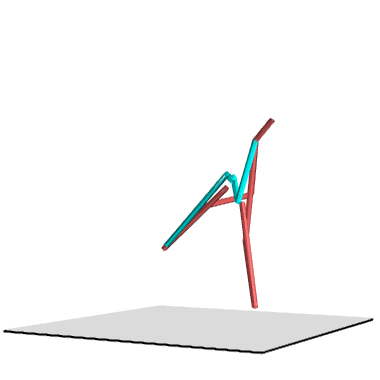}&
 \includegraphics[width=0.08\linewidth]{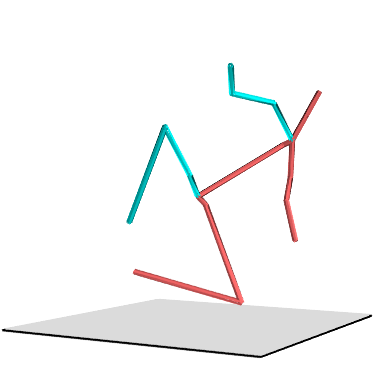}&
 \includegraphics[width=0.08\linewidth]{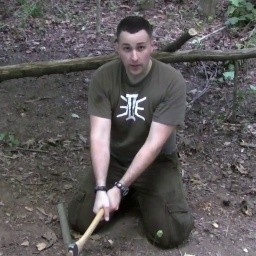}&
 \includegraphics[width=0.08\linewidth]{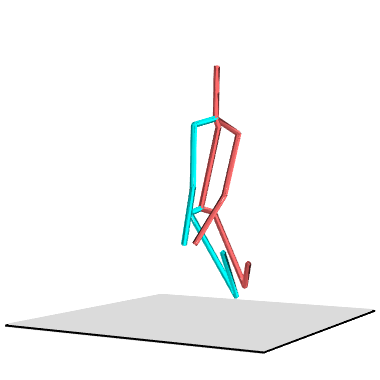}&
 \includegraphics[width=0.08\linewidth]{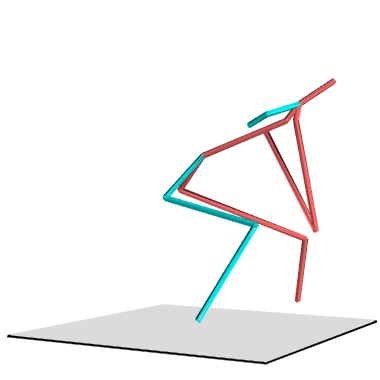}&
 \includegraphics[width=0.08\linewidth]{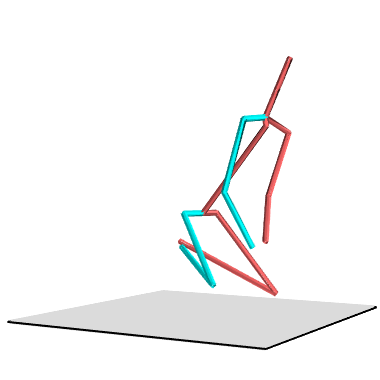}\\
 
 \includegraphics[width=0.08\linewidth]{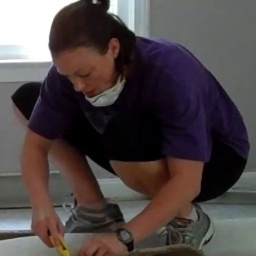}&
 \includegraphics[width=0.08\linewidth]{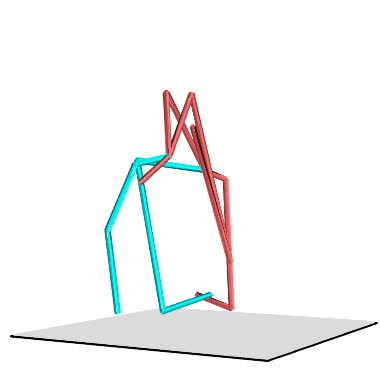}&
 \includegraphics[width=0.08\linewidth]{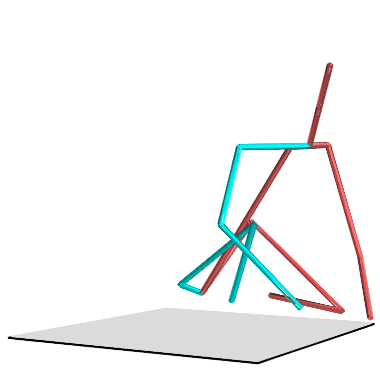}&
 \includegraphics[width=0.08\linewidth]{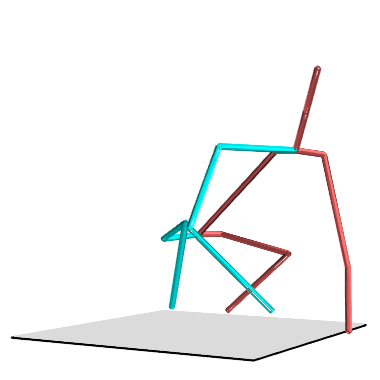}&
 \includegraphics[width=0.08\linewidth]{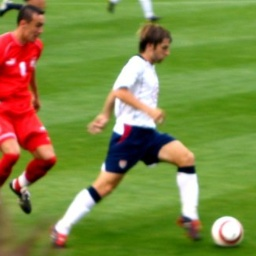}&
 \includegraphics[width=0.08\linewidth]{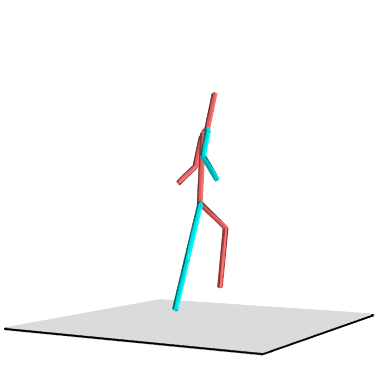}&
 \includegraphics[width=0.08\linewidth]{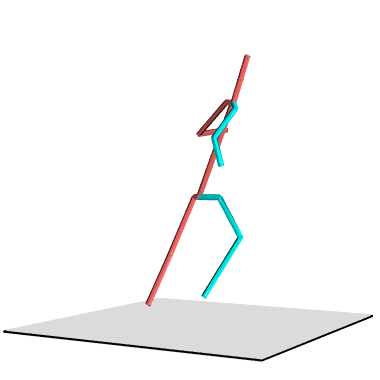}&
 \includegraphics[width=0.08\linewidth]{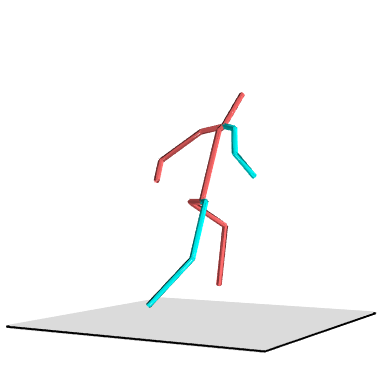}&
 \includegraphics[width=0.08\linewidth]{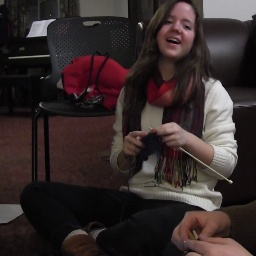}&
 \includegraphics[width=0.08\linewidth]{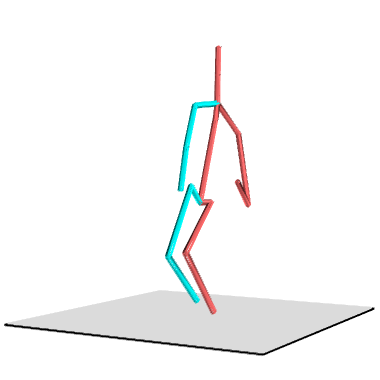}&
 \includegraphics[width=0.08\linewidth]{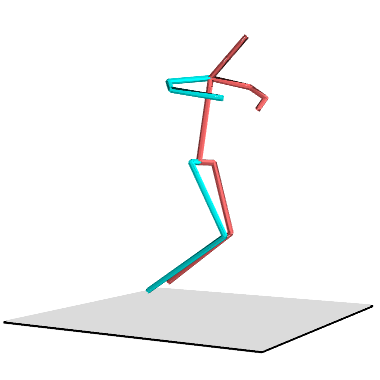}&
 \includegraphics[width=0.08\linewidth]{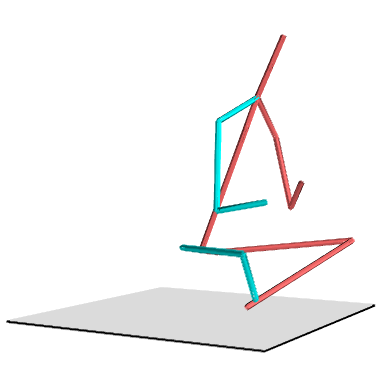}\\

 \includegraphics[width=0.08\linewidth]{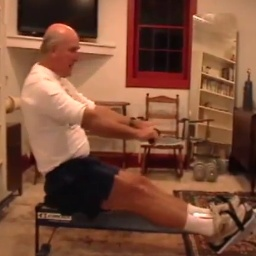}&
 \includegraphics[width=0.08\linewidth]{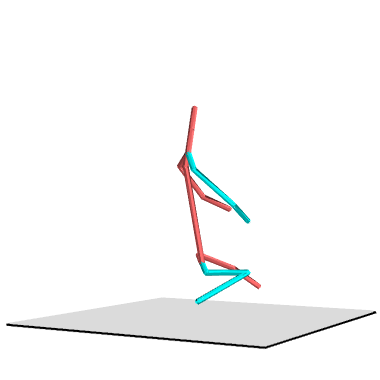}&
 \includegraphics[width=0.08\linewidth]{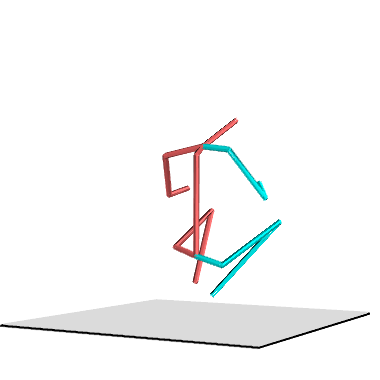}&
 \includegraphics[width=0.08\linewidth]{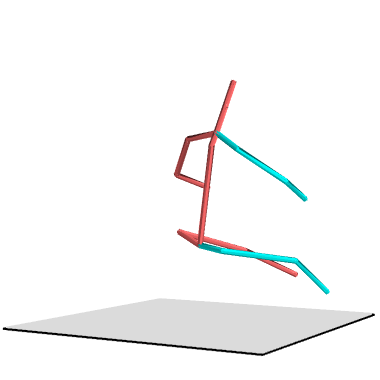}&
 \includegraphics[width=0.08\linewidth]{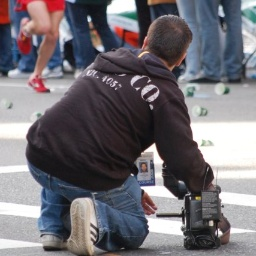}&
 \includegraphics[width=0.08\linewidth]{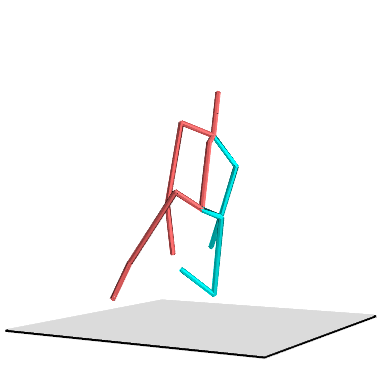}&
 \includegraphics[width=0.08\linewidth]{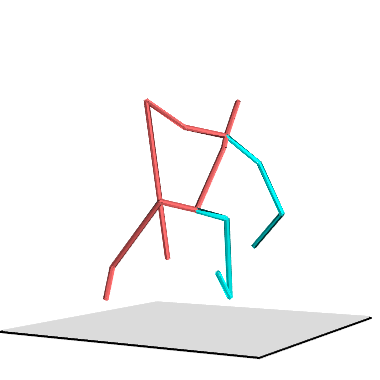}&
 \includegraphics[width=0.08\linewidth]{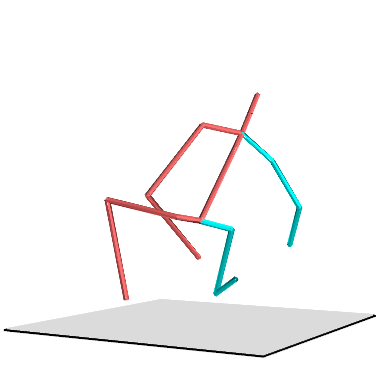}&
 \includegraphics[width=0.08\linewidth]{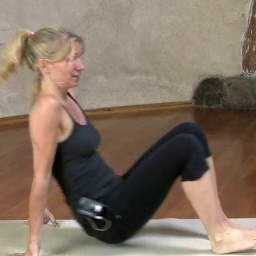}&
 \includegraphics[width=0.08\linewidth]{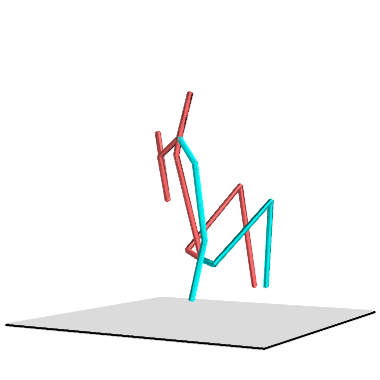}&
 \includegraphics[width=0.08\linewidth]{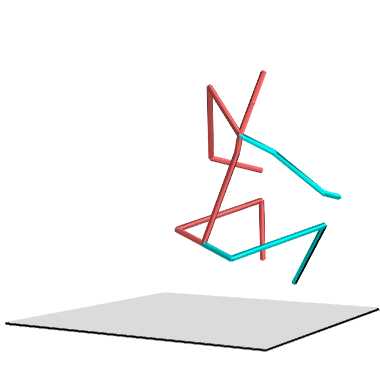}&
 \includegraphics[width=0.08\linewidth]{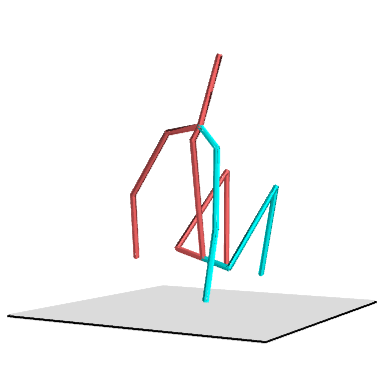}\\

 \includegraphics[width=0.08\linewidth]{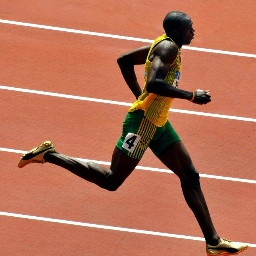}&
 \includegraphics[width=0.08\linewidth]{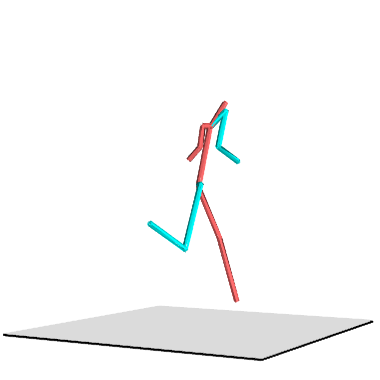}&
 \includegraphics[width=0.08\linewidth]{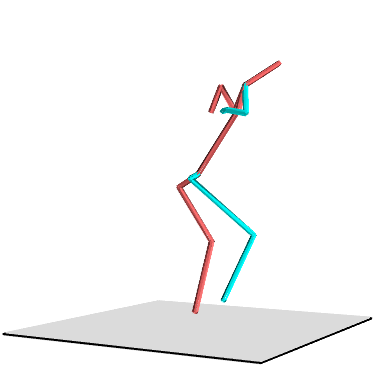}&
 \includegraphics[width=0.08\linewidth]{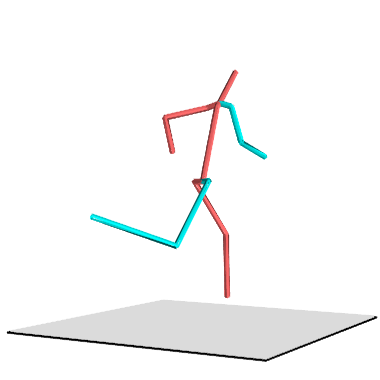}&
 \includegraphics[width=0.08\linewidth]{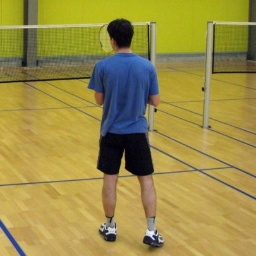}&
 \includegraphics[width=0.08\linewidth]{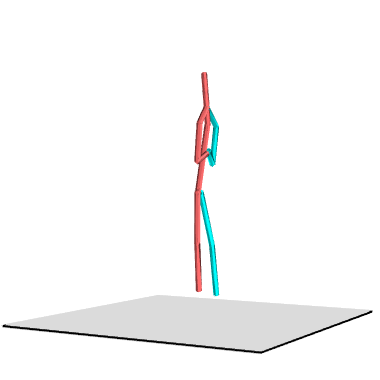}&
 \includegraphics[width=0.08\linewidth]{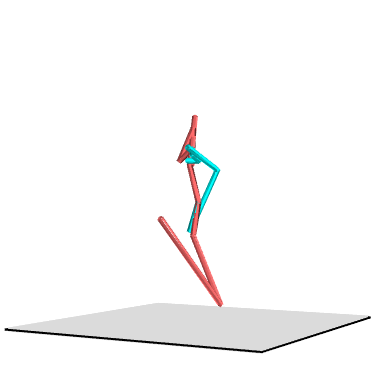}&
 \includegraphics[width=0.08\linewidth]{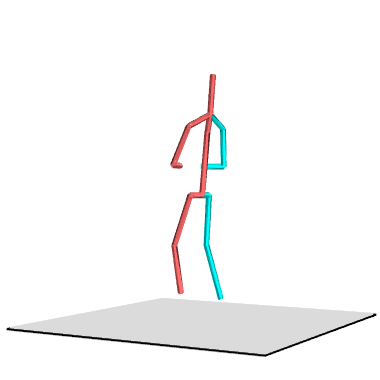}&
 \includegraphics[width=0.08\linewidth]{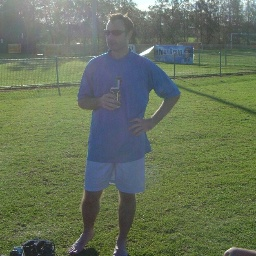}&
 \includegraphics[width=0.08\linewidth]{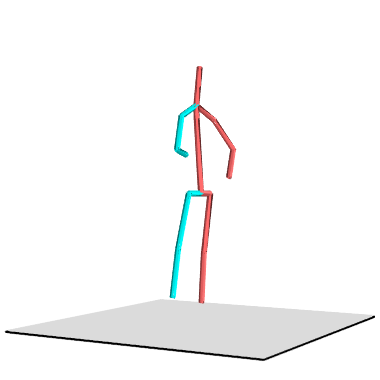}&
 \includegraphics[width=0.08\linewidth]{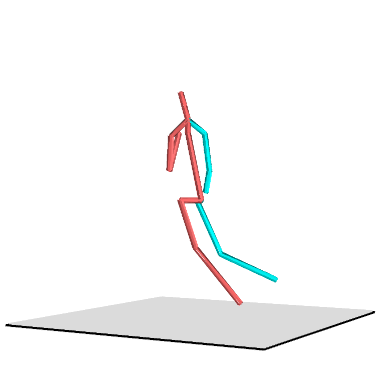}&
 \includegraphics[width=0.08\linewidth]{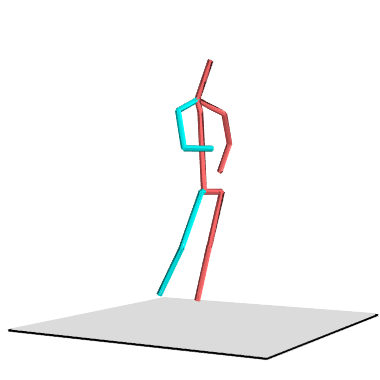}\\
 
 \includegraphics[width=0.08\linewidth]{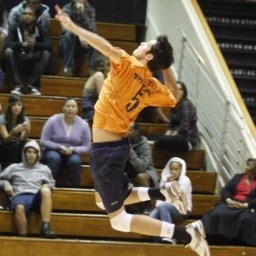}&
 \includegraphics[width=0.08\linewidth]{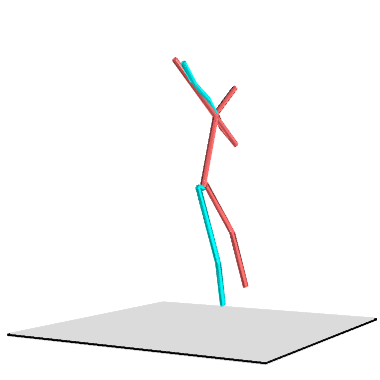}&
 \includegraphics[width=0.08\linewidth]{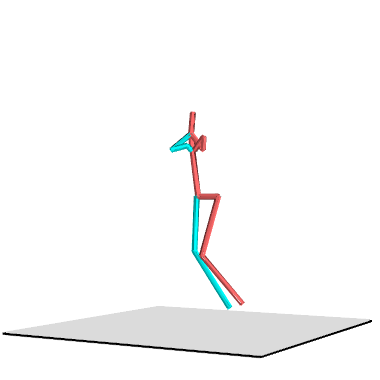}&
 \includegraphics[width=0.08\linewidth]{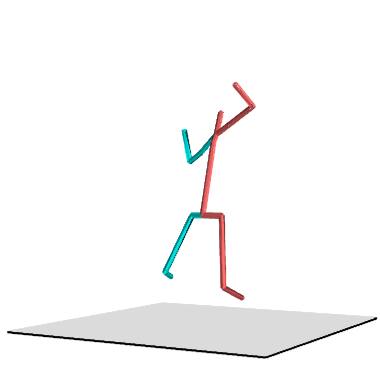}&
 \includegraphics[width=0.08\linewidth]{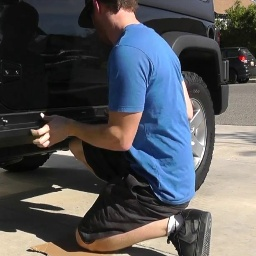}&
 \includegraphics[width=0.08\linewidth]{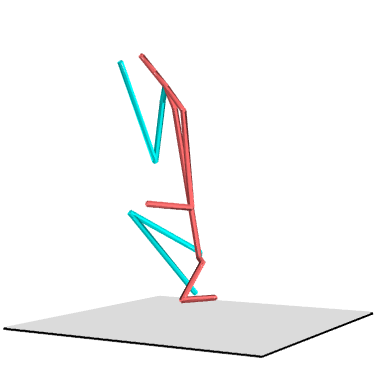}&
 \includegraphics[width=0.08\linewidth]{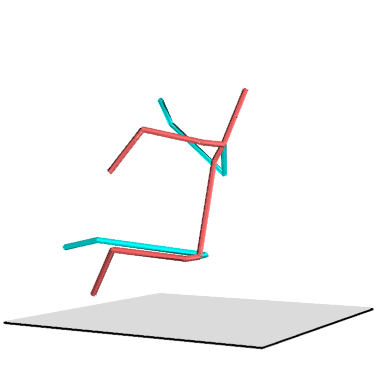}&
 \includegraphics[width=0.08\linewidth]{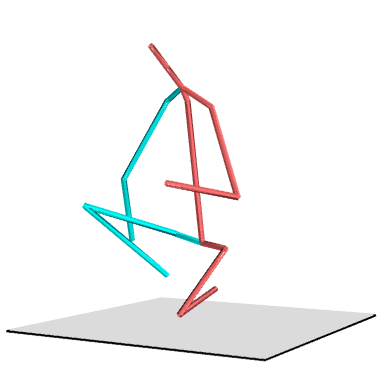}&
 \includegraphics[width=0.08\linewidth]{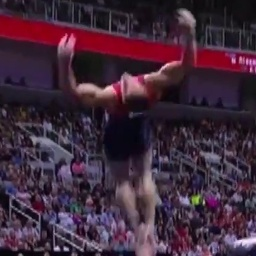}&
 \includegraphics[width=0.08\linewidth]{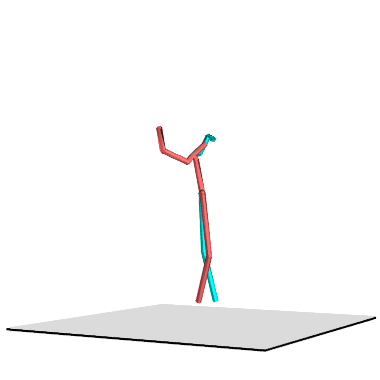}&
 \includegraphics[width=0.08\linewidth]{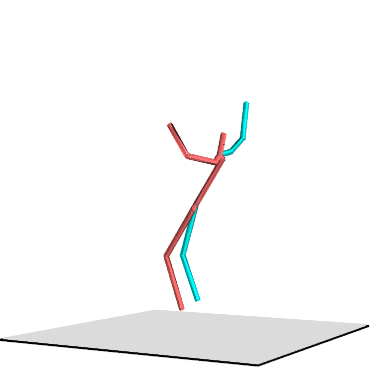}&
 \includegraphics[width=0.08\linewidth]{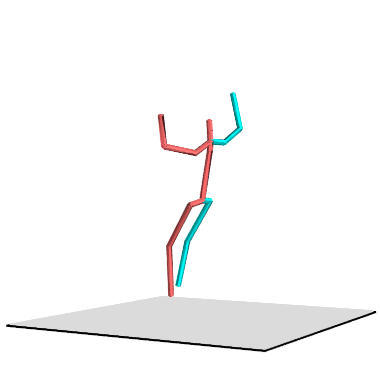}\\

 (a)& (b)& (c)& (d)& 
 (a)& (b)& (c)& (d)& 
 (a)& (b)& (c)& (d)\\

\end{tabular}}
\end{center}
\vspace{-2mm}
\caption{Visual results on the in-the-wild images. The proposed dataset helps to generate more reasonable results in terms of 3D human skeletons. (a) Original in-the-wild images, (b) Results with geometric loss~\cite{zhou2017towards}, (c) Results with GANs~\cite{yang20183d}, (d) Results with the proposed dataset. 3D human pose presents in a novel viewpoint.}
\vspace{-4mm}
\label{vis3}
\end{figure*}
\paragraph{Quantitative Results.}
Table~\ref{exp:pro1} denotes the comparisons with previous methods on the Human3.6M~\cite{h36m_pami}.
The 3D label generator achieves state-of-the-art performance.
For protocol\#1, the generator trained with 2D/3D ground truth from Human3.6M has 17\% (37.6mm vs. 45.5mm) improvements compared with the method of Martinez~\etal~\cite{martinez2017simple}.
To improve the generalization ability, we also train the network with synthetic 2D/3D pairs generated by the unity toolbox.
There is 10\% improvement compared with the method of Martinez~\etal~\cite{martinez2017simple}.
Since the domain gap between the synthetic data and the real data, the model trained with both dataset performs slight worse than the model trained with only the Human3.6M dataset.
However, the qualitative performance on the in-the-wild images is significantly improved, as we will show in the following.
%
%
\paragraph{Qualitative Results.}\vspace{-3mm}
Compared with the method of Martinez~\etal~\cite{martinez2017simple}, we demonstrate the generalization ability qualitatively on the images from MPII and LSP.
Both of the networks to estimate the 3D human pose are based on the 2D ground truth of these datasets, 
As shown in Figure~\ref{exp:intro} (b), (d), the generalization ability of our algorithm outperform the generalization results of Martinez~\etal~\cite{martinez2017simple}.
%
%
Because of the 2D/3D key-points pairs generated by the unity toolbox, the generalization ability of the network is highly improved.
%
\paragraph{Validation of Stereoscopic View Synthesis Subnetwork.}\vspace{-3mm}
To evaluate the quality of synthetic right-view 2D pose, we apply the PCKh metrics following 2D human pose estimation method~\cite{newell2016stacked}.
%
The subnetwork trained with merely Human3.6M achieves 98.2\%, while that trained with the Human3.6M and synthesized dataset by unity toolbox obtains 95.3\% in PCKh-0.5 scores.
It demonstrates that the stereoscopic view synthesis is able to generate high quality right-view 2D pose based on the left-view 2D pose.

\paragraph{Validation of 3D Pose Reconstruction Subnetwork.}\vspace{-3mm}
In Section~\ref{meth:ple}, we note that the stereoscopic architecture can alleviate the depth ambiguity in 3D human pose estimation.
As shown in Table~\ref{exp:pro1}, compared with the monocular structure of Martinez~\etal~\cite{martinez2017simple}, our network without the geometric search scheme has 7.7\% (42.0mm vs. 45.5mm) improvements.
Both of the networks have the same architecture and parameters, the only difference is that we take the 2D poses from multi-view as inputs.
Results illustrate that the rationality of the designed stereoscopic network structure which can boost the performance in 3D human pose estimation.

\paragraph{Validation of the Geometric Search Scheme.}\vspace{-3mm}
We indicate that a high quality 3D human pose can be projected to its 2D counterpart with zero-pixel error.
Based on this premise, we devise a geometric search scheme to further refine the coarse 3D human pose.
As shown in Table~\ref{exp:pro1}, we further analyze the effectiveness of the geometric search scheme by comparing the performance between the method w/o or w/ using it.
The experimental results validate the effectiveness of the geometric search scheme.
%
As one can see in the figure~\ref{exp:intro}(c) and (d), after the geometric search scheme, the predicted 3D human poses become more reasonable.
%
%


\paragraph{Ablation Study}\vspace{-3mm}
We discuss the truth of $\Delta{x}$ in (\ref{equ:map2d}). As shown in Table~\ref{delta}, we conduct experiments on 3D label generator with different $\Delta{x}$. The results show that our generator is not sensitive to $\Delta{x}$. The value of $\Delta{x}$ around 500mm can meet our requirement for generating precise 3D labels. However, $\Delta{x}$ cannot be set with too large value(\eg 5000mm) that will lead to failures in generating 3D poses.

\begin{table}[h]\vspace{-2mm}
\footnotesize
\centering
\caption{Quantitative evaluations on the Human3.6M~\cite{h36m_pami} under Protocol\#1 without using the geometric search scheme and the dataset from the unity toolbox.}\vspace{2mm}
\resizebox{\linewidth}{!}{%
\begin{tabular}{ccccc}
\toprule
$\Delta{x}$/mm& Martinez~\etal~\cite{martinez2017simple}& 250& 500& 750\\ \midrule
Ave./mm& 45.5& 42.2& 42.0& 42.3\\
\bottomrule
\end{tabular}}
\label{delta}
\vspace{-4mm}
\end{table}


\subsection{Evaluations on Baseline Network}
In this section, we mainly compare the two different methods~\cite{zhou2017towards,yang20183d} with the baseline network trained with the proposed in-the-wild 3D pose dataset.
The baseline network is an upgraded version of the Stacked Hourglass Network~\cite{newell2016stacked} with an additional depth regression module.
To focus on the analysis of the proposed 3D pose dataset, we set all the backbone with the same components consisting of 2 stacked hourglass modules, 2 residual blocks, and 2 depth regression modules.
%
%
\paragraph{Quantitative Results.}\vspace{-2mm}
%
%
As shown in Table~\ref{wild_exp}, trained with in-the-wild 3D pose dataset without the additional geometric loss~\cite{zhou2017towards} or the adversarial learning method~\cite{yang20183d}, our baseline network can outperform them.
Compared with Zhou~\etal~\cite{zhou2017towards}, there is about 10.7\% improvement (58.0mm vs. 64.9mm) on the Human3.6M dataset.
It proves that our dataset promotes the accuracy of 3D human estimation on the images taken laboratory scenarios.
\begin{table}[t]
\footnotesize
\centering
\caption{Quantitative evaluations on the Human3.6M~\cite{h36m_pami} under Protocol\#1 (no rigid alignment or similarity transform applied in post-processing). The bold-faced numbers represent the best result.}\vspace{2mm}
\resizebox{\linewidth}{!}{%
\begin{tabular}{lcccc}
\toprule
\textbf{Protocol\#1}                     & Direct. & Discuss   & Eating & Greet   \\ \midrule
Zhou~\etal~\cite{zhou2017towards}        & 54.8    & 60.7      & 58.2   & 71.4    \\
Yang~\etal~\cite{yang20183d}             & 53.0    & 60.8      & \textbf{47.9}   & 57.1    \\
Ours                                     & \textbf{47.4}    & \textbf{56.4}      & 49.4   & \textbf{55.7}    \\ \midrule
                                         & Phone   & Photo     & Pose   & Purch.  \\ \midrule
Zhou~\etal~\cite{zhou2017towards}        & 62.0    & 65.5      & 53.8   & 55.6    \\
Yang~\etal~\cite{yang20183d}             & 61.5    & \textbf{65.5}      & 50.8   & 49.9    \\
Ours                                     & \textbf{58.0}    & 67.3      & \textbf{46.0}   & \textbf{46.0}    \\ \midrule
                                         & Sitting & SittingD. & Smoke  & Wait    \\ \midrule
Zhou~\etal~\cite{zhou2017towards}        & 75.2    & 111.6     & 64.1   & 66.0    \\
Yang~\etal~\cite{yang20183d}             & 73.3    & \textbf{98.6}      & 58.8   & 58.1    \\
Ours                                     & \textbf{67.7}    & 102.4     & \textbf{57.0}   & \textbf{57.3}    \\ \midrule
                                         & WalkD.  & Walk      & WalkT. & Average \\ \midrule
Zhou~\etal~\cite{zhou2017towards}        & 51.4    & 63.2      & 55.3   & 64.9    \\
Yang~\etal~\cite{yang20183d}             & 42.0    & 62.3      & 43.6   & 59.7    \\
Ours                                     & \textbf{41.1}    & \textbf{61.4}      & \textbf{40.7}   & \textbf{58.0}    \\ \bottomrule
\end{tabular}}
\label{wild_exp}
\vspace{-4mm}
\end{table}

\paragraph{Qualitative Results.}\vspace{-2mm}
By visualizing the 3D skeleton of the predicted human body, we show that the model trained with the in-the-wild 3D pose data is robust in the realistic scenes.
Figure~\ref{vis3} shows the visualization results by different methods.
As one can see in the figure, our baseline network trained with the proposed in-the-wild 3D pose dataset can estimate more reasonable results, which in term proves the quality of our proposed dataset.
In addition, we discover that our model can handle challenging samples such as leaning over, sitting cross-legged, and jumping.
\paragraph{Cross-Domain Generalization.}
%
%
We further verify the generalization introduced by our proposed 3D pose dataset on the MPI-INF-3DHP~\cite{mehta2016monocular}.
Without any retraining the model on this dataset, we compare the results by Zhou~\etal~\cite{zhou2017towards}, Yang~\etal~\cite{yang20183d} and our baseline network.
As reported in Table~\ref{exp:mpi-inf}, one can observe that the method trained with in-the-wild 3D data significantly improves the generalization ability.
%

%
%
\begin{table}[t]
\center
\footnotesize
\caption{Quantitative evaluations on the MPI-INF-3DPH~\cite{mehta2017monocular}. No training data from this dataset have been used for training by any method.}\vspace{2mm}
\begin{tabular}{lccc}
\toprule
                          &   Zhou~\etal~\cite{zhou2017towards}  & Yang~\etal~\cite{yang20183d}         & Ours                \\ \midrule
            PCK           &   69.2                               &   69.0                               &\textbf{71.2}        \\  
            AUC           &   32.5                               &   32.0                               &\textbf{33.8}        \\\bottomrule
\label{exp:mpi-inf}
\end{tabular}
\end{table}

\paragraph{Generalization Evaluations by Action Classification.}
%
\begin{figure}[t]
\centering
\includegraphics[width=\linewidth]{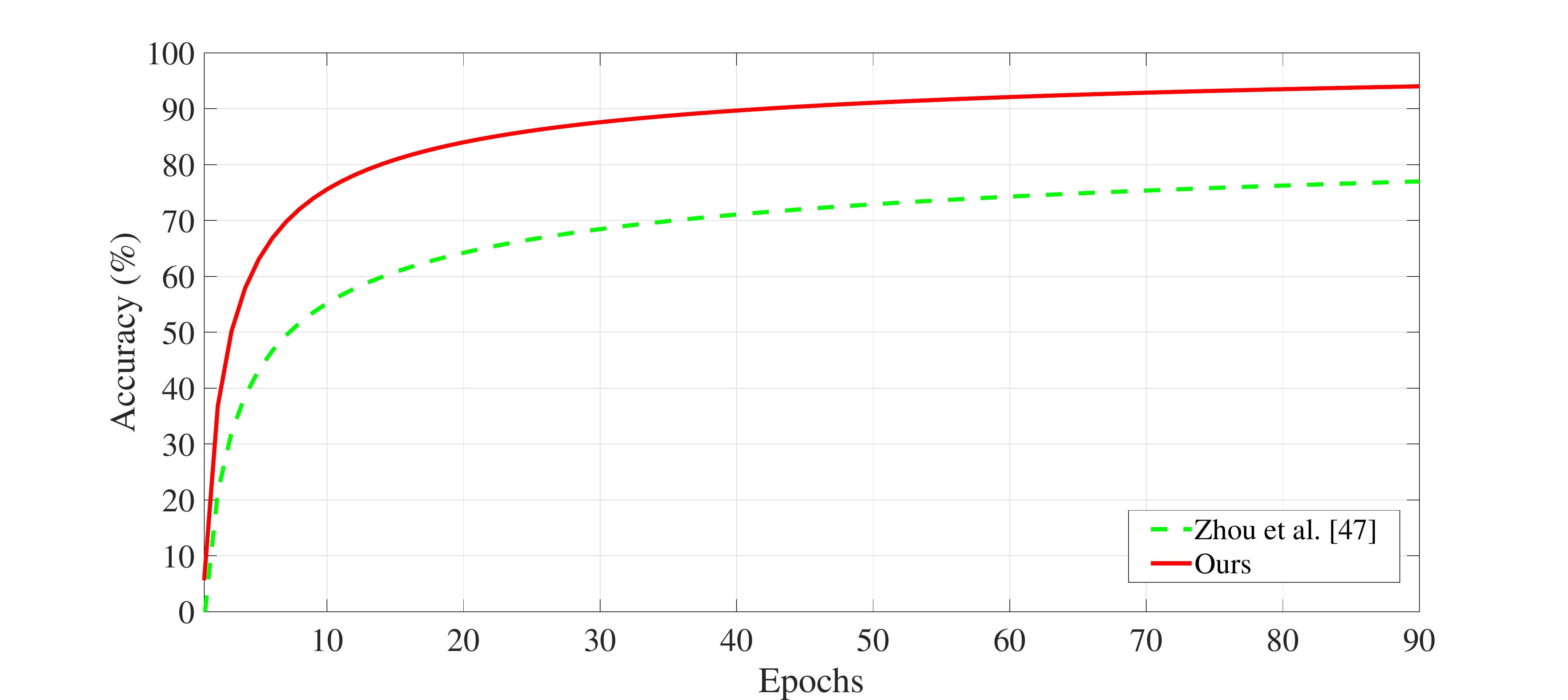}
\caption{Evaluation results of action classification on Penn action dataset~\cite{zhang2013actemes}. The detected 3D human pose by our method is more conducive to the action classification task.}
\label{exp:act_curve}
\vspace{-4mm}
\end{figure}
To further evaluate the generalization ability of different methods, we propose an approach to utilize detected 3D joints for action classification on the Penn Action dataset~\cite{zhang2013actemes}.
%
Using only the coordinate location of predicted 3D joints, we devise a simple network consist of multiple fully connected layers, the detailed network structure can be found in our supplementary file.
%
%
%
When training this network, we use the location of predicted 3D joints of 25 consecutive frames in Penn Action dataset~\cite{zhang2013actemes} as the inputs.
%
%

\begin{figure}[!t]
\footnotesize
\centering
\renewcommand{\tabcolsep}{1pt} 
\renewcommand{\arraystretch}{1} 
\begin{center}
\resizebox{1\linewidth}{!}{%
\begin{tabular}{ccccc}
  \includegraphics[width=0.24\linewidth]{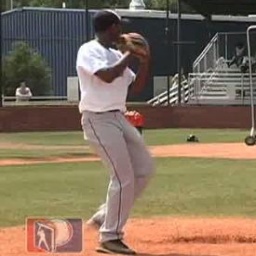} &
  \includegraphics[width=0.24\linewidth]{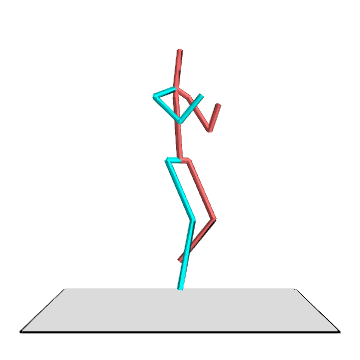} &
  \includegraphics[width=0.24\linewidth]{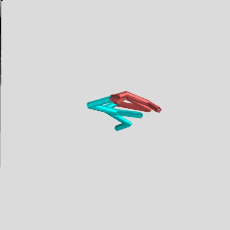} &
  \includegraphics[width=0.24\linewidth]{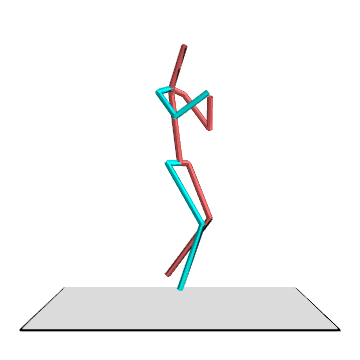} &
  \includegraphics[width=0.24\linewidth]{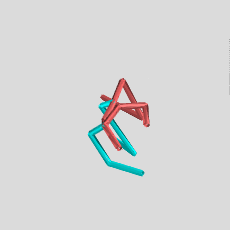} \\

  \includegraphics[width=0.24\linewidth]{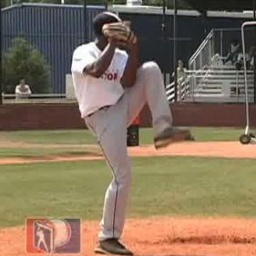} &
  \includegraphics[width=0.24\linewidth]{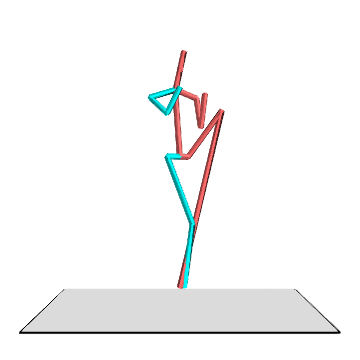} &
  \includegraphics[width=0.24\linewidth]{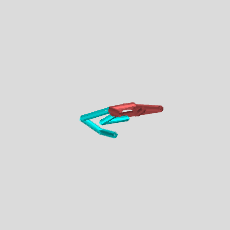} &
  \includegraphics[width=0.24\linewidth]{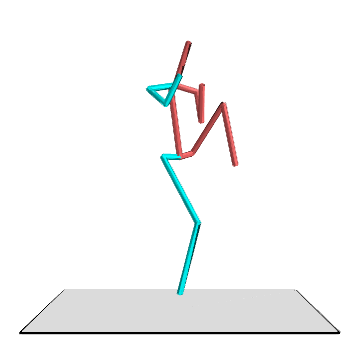} &
  \includegraphics[width=0.24\linewidth]{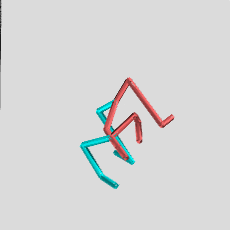} \\

  \includegraphics[width=0.24\linewidth]{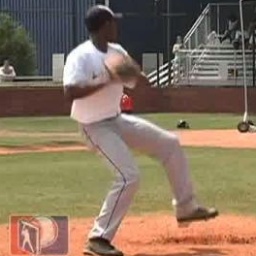} &
  \includegraphics[width=0.24\linewidth]{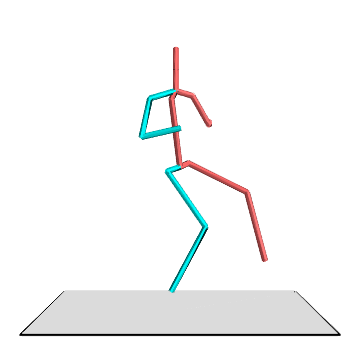} &
  \includegraphics[width=0.24\linewidth]{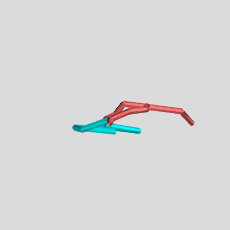} &
  \includegraphics[width=0.24\linewidth]{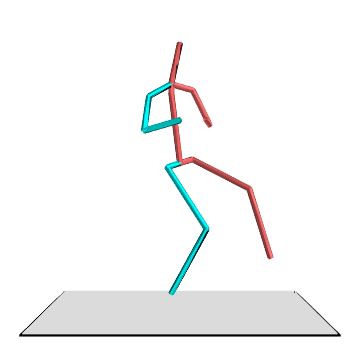} &
  \includegraphics[width=0.24\linewidth]{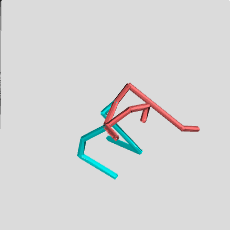} \\

  \includegraphics[width=0.24\linewidth]{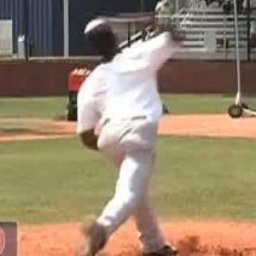} &
  \includegraphics[width=0.24\linewidth]{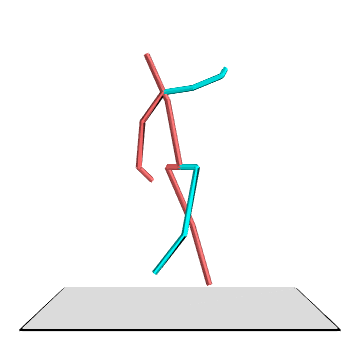} &
  \includegraphics[width=0.24\linewidth]{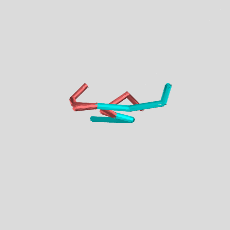} &
  \includegraphics[width=0.24\linewidth]{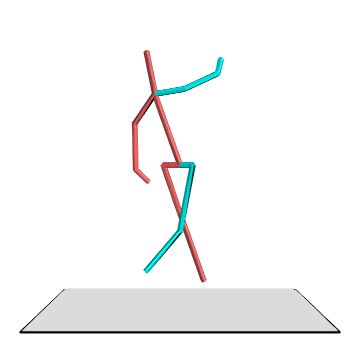} &
  \includegraphics[width=0.24\linewidth]{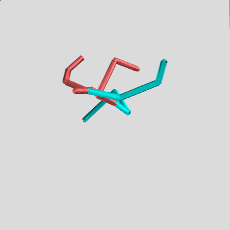} \\
  (a)& (b)& (c)& (d)& (e)\\
    \end{tabular}}
\end{center}
\vspace{-2mm}
\caption{Visualization of the 3D skeletons of the video sequence extracted in the Penn~\cite{zhang2013actemes} dataset. (a) Video sequences, (b) 3D human skeletons from the front viewpoint predicted by Zhou~\etal~\cite{zhou2017towards}, (c) 3D human skeletons from the top viewpoint predicted by Zhou~\etal~\cite{zhou2017towards}, (d) 3D human skeletons from the front viewpoint predicted by our method, (e) 3D human skeletons from the top viewpoint predicted by our method.}
\vspace{-2mm}
\label{exp:video}
\end{figure}

As shown in Figure~\ref{exp:act_curve}, we can find that the model trained with our predicted 3D joints are more precise than the model of Zhou \etal~\cite{zhou2017towards} (93\% vs. 80\%).
In addition, we analyze the difference between the detected 3D joints of the two models.
%
%
Figure~\ref{exp:video} shows an example that the method of Zhou~\etal~\cite{zhou2017towards} predicts 3D pose at almost the same depth level, while the depth of the predicted 3D poses by our baseline network varies with the baseball player movements.
The results demonstrate that the inaccurate predicted depth leads to a lower accuracy of action classification. 
%
The superior performance in action recognition task can further prove the generalization of our baseline network.


\section{Conclusions}
In this paper, we solve the generalization problem of 3D human pose estimation from a novel perspective.
We propose a principled approach to generate high quality 3D labels given an in-the-wild image automatically.
Based on the stereo inspired structure, the proposed network with a carefully designed geometric search scheme significantly outperforms other methods quantitatively and qualitatively.
We proposed an in-the-wild 3D pose dataset containing more than 400,000 images by employing this network as a 3D label generator.
%
%
Experimental results show that the baseline model trained with the proposed dataset can significantly improve the performance on the public Human3.6M and boost the generalization ability on the in-the-wild images.
%
%
In the future work, we plan to investigate to generate high quality 3D labels on the in-the-wild videos.
\clearpage
{\small
\bibliographystyle{ieee}
\bibliography{Mono3DPose}
}

\end{document}

%% file: macro.tex
\usepackage{times}
\usepackage{epsfig}
\usepackage{xspace}
\usepackage{color}
\usepackage{multirow}
\usepackage{graphics}
\usepackage{adjustbox}
\usepackage{subfigure}
\usepackage{amsmath}
\usepackage[noend]{algpseudocode}
\usepackage{algorithmicx,algorithm}
\usepackage{amsfonts,amssymb}
\usepackage{paralist} 
\usepackage{enumitem}
\usepackage{graphicx} 
\usepackage{epstopdf}
\usepackage{booktabs} 
\usepackage[noend]{algpseudocode}
\usepackage{algorithmicx,algorithm}

\usepackage[pagebackref=true,breaklinks=true,letterpaper=true,colorlinks,citecolor=blue,linkcolor=blue,bookmarks=false]{hyperref}

\long\def\ignorethis#1{}

\definecolor{Gray}{rgb}{0.35,0.35,0.35}
\definecolor{Blue}{rgb}{0,0.2,0.8}
\definecolor{Red}{rgb}{0.8,0.2,0}
\definecolor{Green}{rgb}{0.0,0.5,0.1}
\definecolor{Gray}{rgb}{0.4,0.4,0.4}

\newlength\paramargin
\newlength\figmargin
\newlength\secmargin

\usepackage{array}
\newcolumntype{L}[1]{>{\raggedright\let\newline\\\arraybackslash\hspace{0pt}}m{#1}}
\newcolumntype{C}[1]{>{\centering\let\newline\\\arraybackslash\hspace{0pt}}m{#1}}
\newcolumntype{R}[1]{>{\raggedleft\let\newline\\\arraybackslash\hspace{0pt}}m{#1}}

\def\ie{i.e.,~}
\def\eg{e.g.,~}

\def\etal{et~al.\xspace}

\graphicspath{{figure/eps/}}